\documentclass[lettersize,journal]{IEEEtran}
\usepackage{amsmath,amsfonts}
\usepackage{algorithmic}
\usepackage{algorithm}
\usepackage{array}
\usepackage[caption=false,font=normalsize,labelfont=sf,textfont=sf]{subfig}
\usepackage{textcomp}
\usepackage{stfloats}
\usepackage{url}
\usepackage{verbatim}
\usepackage{graphicx}
\usepackage{cite}
\usepackage{booktabs} 
\usepackage{multirow}
\usepackage{siunitx}
\usepackage{longtable}
\usepackage[normalem]{ulem}
\begin{document}


\title{GAIA: A Global, Multi-modal, Multi-scale Vision-Language Dataset for Remote Sensing Image Analysis}

\author{Angelos Zavras, Dimitrios Michail, Xiao Xiang Zhu~\IEEEmembership{Fellow Member, IEEE}, Beg{\"u}m Demir~\IEEEmembership{Senior Member, IEEE}, Ioannis Papoutsis
\thanks{Angelos Zavras is with the Orion Lab, School of Rural, Surveying and Geoinformatics Engineering, National Technical University of Athens, 15772 Athens, Greece \& Institute of Astronomy, Astrophysics, Space Applications and Remote Sensing, National Observatory of Athens, 11810 Athens, Greece, and with the Department of Informatics and Telematics, Harokopio University of Athens, 17676 Athens, Greece (e-mail: azavras@noa.gr).}
\thanks{Dimitrios Michail is with the Department of Informatics and Telematics, Harokopio University of Athens, 17676 Athens, Greece (e-mail: michail@hua.gr).}
\thanks{Xiao Xiang Zhu is with the Chair of Data Science in Earth Observation, Technical University of Munich, 80333 Munich, Germany, and also with the Munich Center for Machine Learning, 80333 Munich, Germany (e-mail: xiaoxiang.zhu@tum.de).}
\thanks{Beg\"um Demir is with the Faculty of Electrical Engineering and Computer Science, Technische Universit\"at Berlin, 10623 Berlin, Germany, also with the BIFOLD - Berlin Institute for the Foundations of Learning and Data, 10623 Berlin, Germany (e-mail: demir@tu-berlin.de).}
\thanks{Ioannis Papoutsis is with the Orion Lab, School of Rural, Surveying and Geoinformatics Engineering, National Technical University of Athens, 15772 Athens, Greece \& Institute of Astronomy, Astrophysics, Space Applications and Remote Sensing, National Observatory of Athens, 11810 Athens, Greece (e-mail: ipapoutsis@mail.ntua.gr).}}

\maketitle

\begin{abstract}
The continuous operation of Earth-orbiting satellites generates vast and ever-growing archives of Remote Sensing (RS) images. Natural language presents an intuitive interface for accessing, querying, and interpreting the data from such archives. However, existing Vision-Language Models (VLMs) are predominantly trained on web-scraped, noisy image-text data, exhibiting limited exposure to the specialized domain of RS. This deficiency results in poor performance on RS-specific tasks, as commonly used datasets often lack detailed, scientifically accurate textual descriptions and instead emphasize solely on attributes like date and location. To bridge this critical gap, we introduce GAIA, a novel dataset designed for multi-scale, multi-sensor, and multi-modal RS image analysis. GAIA comprises of 201,005 meticulously curated RS image-text pairs, representing a diverse range of RS modalities associated to different spatial resolutions. Unlike existing vision-language datasets in RS, GAIA specifically focuses on capturing a diverse range of RS applications, providing unique information about environmental changes, natural disasters, and various other dynamic phenomena. The dataset provides a spatially and temporally balanced distribution, spanning across the globe, covering the last 25 years with a balanced temporal distribution of observations. GAIA's construction involved a two-stage process: (1) targeted web-scraping of images and accompanying text from reputable RS-related sources, and (2) generation of five high-quality, scientifically grounded synthetic captions for each image using carefully crafted prompts that leverage the advanced vision-language capabilities of GPT-4o. We also release an automated processing framework developed for this purpose, enabling the broader research community to generate captions for RS images using the web-crawled image-text data. Our extensive experiments, including fine-tuning of CLIP and BLIP2 models, demonstrate that GAIA significantly improves performance on RS image classification, cross-modal retrieval and image captioning tasks, proving its value as a crucial resource for advancing the field. We make our dataset, automated processing framework and fine-tuned model weights publicly available on our project's GitHub repository\footnote{\url{https://github.com/Orion-AI-Lab/GAIA}}.
\end{abstract}


\section{Introduction}
\IEEEPARstart{T}{he} Earth's orbit is teeming with a growing constellation of Earth Observation (EO) satellites, continuously producing an unprecedented volume of diverse and complex information about our planet. According to the latest Copernicus Sentinel Data Access Annual Report~\footnote{\url{https://sentinels.copernicus.eu/web/sentinel/-/ninth-copernicus-sentinel-data-access-annual-report}}, Europe's Copernicus satellite constellation alone features an average daily publication rate of more than 20 TiB of user-level data. More specifically, within the reporting period from January 1, 2023 to October 31, 2023, a total of more than 11.5 million user-level data were published, accounting for a total data volume of 6.04 PiB. To put this into perspective, the 6.04 PiB published during the aforementioned reporting period alone is more than ESA’s entire collection of EO data from the pre-Copernicus era, which amounted to 5.6 PB at the end of 2013. It is evident that this exponential growth in data production has outpaced our capacity for meaningful information extraction, prompting the Remote Sensing (RS) community to adopt Deep Learning (DL) approaches to extract meaningful insights for Earth observation.

In this data-rich landscape~\cite{xiong2024earthnets}, language provides a natural interface to interact with and analyze these vast RS archives. Language provides a compressed representation of reality, efficiently encoding concepts. However, this compression is lossy and language alone lacks vital knowledge of the physical world that can only be acquired through sensory experience. To this end, Vision-Language models (VLMs) offer a solution by leveraging the richness of visual (image) data alongside the compression of textual information. By combining these modalities, VLMs are able to learn abstract representations that encapsulate comprehensive scene understanding. VLMs hold immense potential for the RS domain. By comprehending concepts from both visual and language modalities, VLMs can enable intuitive interfaces for RS data analysis (e.g. image captioning, cross-modal retrieval and visual question answering) or even serve as versatile building blocks for next-generation of RS foundation models. 

Data is regarded as the cornerstone for VLM training in the computer vision domain, a principle firmly established in the foundation models literature~\cite{bommasani2021opportunities}, and the RS domain is no exception. Existing general purpose VLMs are trained on large amounts of web-crawled data, to encode general knowledge, without emphasizing on specialized domains such as RS and medical imagery, which as a matter of fact demonstrate fundamentally different distributions compared to natural images encountered during pre-training. Perhaps most critically, unlike natural image-text datasets, which can easily reach billion-scale~\cite{schuhmann2022laion, gadre2024datacomp}, specialized datasets hold a relatively limited scale due to their innate demand for highly targeted knowledge and substantial human labor, rendering it costly and often impractical. 

The advancements in VLMs have drawn significant attention in the RS field, leading to studies across different applications. While early attempts have shown the potential of applying VLMs to RS, it is still an emerging field with many unsolved challenges~\cite{li2024vision}. A challenging issue that impedes the development of VLMs in RS is the lack of large-scale aligned image-text datasets. Existing RSI datasets mostly focus on visual recognition tasks and do not provide language annotations. While there have been some efforts to create RS image-text paired datasets, their scale, quality, and diversity still leave room for improvement.

Recent initiatives such as LAION-EO~\cite{czerkawski2023laion} have demonstrated the presence of RS-related samples within large-scale web-crawled image-text paired datasets like LAION-5B~\cite{schuhmann2022laion}. However, these samples frequently lack the depth and specificity needed for domain-specific RS applications. They primarily focus or are limited to basic attributes such as date and location, neglecting critical RS-specific semantic information including land cover types, geospatial context, and environmental features. This limits the applicability of these datasets for developing high-quality RS image analysis models, which require precise and context-rich in-domain annotations.

Existing datasets for training VLMs in the RS domain suffer from significant limitations that hinder the development of effective in-domain models or the adaptation of existing pre-trained VLMs to RS tasks. To begin with, the predominant reliance on aerial and Very High-Resolution (VHR) commercial satellite imagery in widely used, yet limited-scale, datasets~\cite{qu2016deep, lu2017exploring, yuan2022exploring, zhan2023rsvg} introduces two primary challenges. First, the proprietary nature of this data restricts accessibility and reproducibility, contrasting with the open-access paradigm promoted by missions like the Copernicus Sentinel. Second, the finer spatial resolution inherent in these datasets, while advantageous for specific applications, creates a domain shift compared to data from non-commercial, often coarser-resolution satellite missions. On the other hand, more recent works exploit existing class-labeled datasets in order to automatically generate synthetic captions reaching up to million-scale capacity. Despite addressing the dataset scale aspect, these methods, similar to existing manually curated datasets, fail to adequately support the development of robust in-domain VLMs capable of generalizing across the diverse landscape of RS applications and data sources, due to their severely limited level of caption detail and lack of diverse, domain-specific vocabulary.

To address these challenges, we release GAIA, a dataset of approximately 40,201 RS images, each accompanied by five diverse and informative captions, resulting in 201,005 image-text pairs. Specifically, less that 3\% of GAIA's images are found in compared web-crawled datasets. Moreover, we provide extensive benchmarks for RS image classification and cross-modal retrieval tasks by fine-tuning CLIP on the GAIA dataset, and we publish the pre-trained model weights to facilitate further research. Complementing the dataset, we introduce a novel automated processing framework for generating RS image descriptions by leveraging web-crawled data from reputable sources and the advanced language generation capabilities of ChatGPT. Our work aims to bridge the gap between the vast potential of VLMs and the unique challenges of the RS domain, accelerating the development of foundation models that can unlock new insights from Earth observation data. We make our dataset, automated processing framework and fine-tuned model weights publicly available on our project's GitHub repository\footnote{\url{https://github.com/Orion-AI-Lab/GAIA}}.

Our main contributions can be summarized as follows:
\begin{enumerate}
    \item We introduce GAIA, a carefully curated RS dataset with 40,201 images along 5 synthetic captions each (201,005 image-text pairs) featuring global, multi-modal, multi-scale coverage spanning across the last 25 years. It exhibits minimal overlap ($<3\%$) with existing datasets, enabling diverse Earth event analysis across multiple spatial resolutions and modalities.
    \item We release a novel, automated processing framework for generating detailed and scientifically grounded textual descriptions and rich metadata for RS imagery. Our framework leverages targeted web-scraped data from reputable RS websites and harnesses the advanced vision-language capabilities of GPT-4o to produce five high-quality captions per image, significantly enriching the semantic content compared to original alt-text. 
    \item We evaluate GAIA’s impact extensively by fine-tuning CLIP and BLIP2 models on both alt-text and synthetic captions for RS scene classification, cross-modal retrieval, and image captioning. Our benchmarking demonstrates substantial performance improvements, with synthetic captions proving superior in enhancing semantic alignment and task accuracy. We release pre-trained models to facilitate future RS VLM-related research.
\end{enumerate}

\section{Related work}
VLMs have emerged as powerful tools for understanding and generating connections between visual and textual information. Pioneering works like CLIP~\cite{radford2021learning} demonstrated the effectiveness of contrastive learning in creating aligned visual-semantic embeddings, while more recent models such as LLaVA~\cite{liu2024visual,liu2024improved} have expanded these capabilities to include sophisticated visual reasoning and open-ended dialogue. Modern VLMs can perform a wide range of tasks, including image captioning, visual question answering, and cross-modal retrieval without task-specific fine-tuning. The success of these models stems from their ability to learn generalizable visual-semantic representations through self-supervised learning on vast amounts of image-text data, typically utilizing web-scale datasets containing billions of image-text pairs. This approach has proven particularly effective at capturing nuanced relationships between visual content and natural language descriptions, enabling zero-shot transfer to novel tasks and domains.

\subsection{Have existing VLMs seen Remote Sensing imagery?} \label{sec:seen_rs}
While VLMs excel at general visual and textual understanding, their training on massive, web-scraped datasets like LAION-5B raises concerns about their proficiency in specialized domains such as RS~\cite{panigrahi2023have}. Recent studies employing rigorous filtering techniques on these large-scale datasets have provided compelling evidence that VLMs have indeed encountered RS images, but this exposure is incidental, unsystematic, and ultimately superficial. This limited and unfocused engagement with RS data highlights a crucial gap in the capabilities of current VLMs.

\begin{figure*}
\centering
\includegraphics[width=\textwidth]{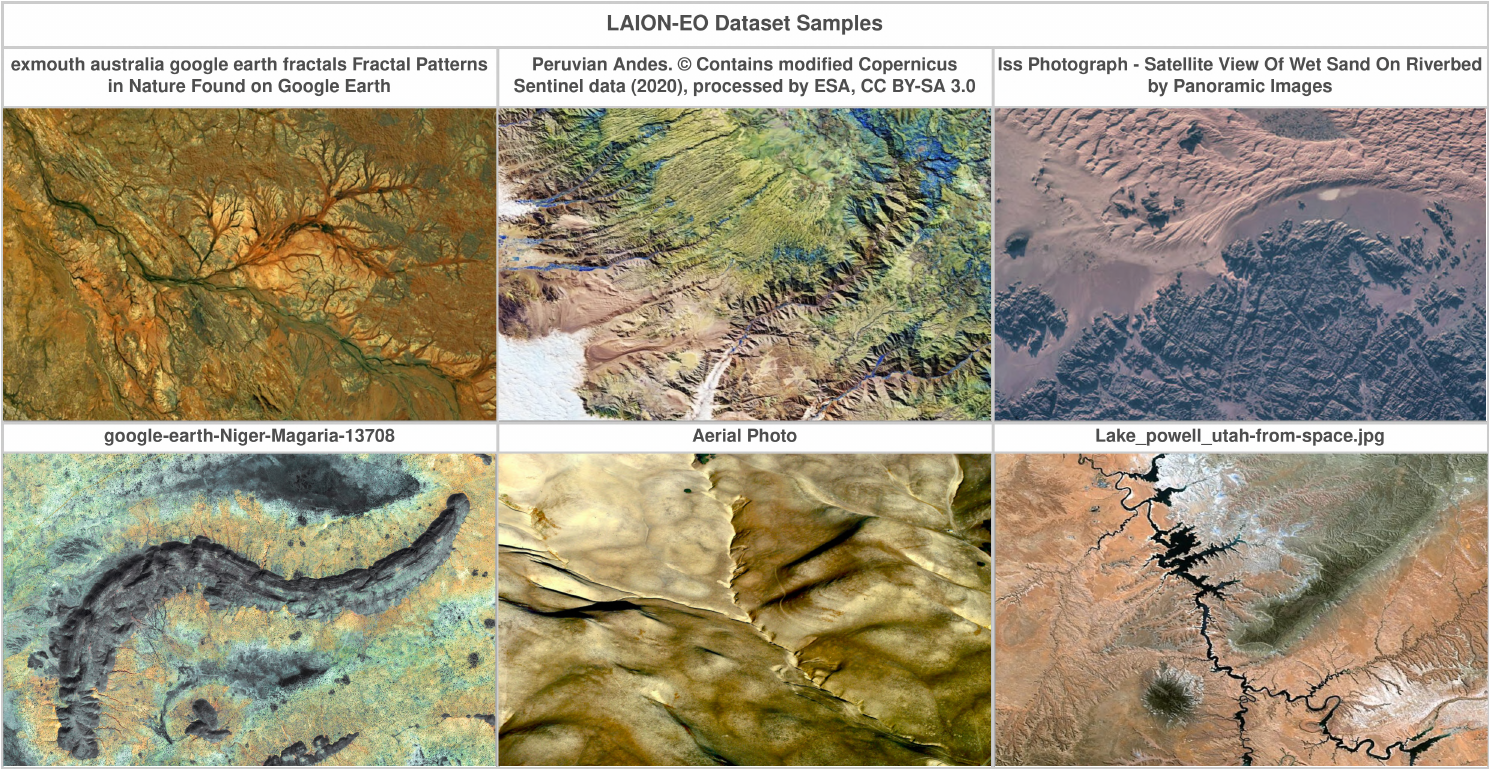}
\caption{Representative samples from the LAION-EO~\cite{czerkawski2023laion} dataset illustrate a fundamental limitation inherent in web-scraped image-text paired datasets for remote sensing: despite images exhibit sufficient visual fidelity, the accompanying textual descriptions are characterized by high noise levels and lack domain-specific details, diminishing their utility for Earth Observation tasks.}
\label{fig:laion_samples}
\end{figure*}

Several research efforts have focused on extracting RS subsets from web-crawled large-scale general-purpose datasets, extensively used within the computer vision domain. As an example, Czerkawski and Francis developed LAION-EO~\cite{czerkawski2023laion}, leveraging CloudSEN12~\cite{aybar2022cloudsen12} as an "anchor dataset" of Sentinel-2 images to retrieve similar images from LAION-5B. Their multi-stage filtering process, involving nearest neighbor search and refined CLIP-based similarity thresholds, yielded 24k samples in the prototype version and an expanded set of 113k in version 1. Similarly, LAION-RS~\cite{wang2024skyscript} identified a subset of 726k RS images, i.e. a mere 0.03\% of LAION-2B, using a binary classifier. Zhang et. al.~\cite{zhang2024rs5m} introduced RS5M, a dataset of 5 million RS image-text pairs. A core component of RS5M, the PUB11 subset containing over 3 million pairs, was created by filtering 11 large-scale public datasets using keyword-based filtering, de-duplication, and a VLM-based filter. The 11 filtered large-scale public datasets, include LAION2B-en~\cite{schuhmann2022laion}, LAION400M~\cite{schuhmann2021laion}, LAIONCOCO~\cite{schuhmann2022laioncoco}, COYO700M~\cite{byeon2022coyo}, CC3M~\cite{sharma2018conceptual}, CC12M~\cite{changpinyo2021conceptual}, YFCC15M~\cite{thomee2016yfcc100m}, WIT~\cite{srinivasan2021wit}, Redcaps~\cite{desai2021redcaps}, SBU~\cite{ordonez2011im2text}, and Visual Genome~\cite{krishna2017visual}. These filtering efforts offer critical insights into the current state of VLMs. First, the relatively small size of the extracted subsets, such as the minuscule 0.03\% of LAION-2B forming LAION-RS, demonstrates the limited and incidental exposure of VLMs to RS data. Second, as shown in Fig.~\ref{fig:laion_samples}, web-scraped datasets like LAION-EO present a significant challenge: while the extracted imagery is generally of sufficient quality and resolution, the associated captions are inadequate for RS image analysis. They are dominated by generic attributes and lack the detailed, domain-specific descriptions, thus fail to describe land cover types, specific geographical features, or other essential elements that are crucial in RS downstream tasks.

These findings demonstrate that the current generation of VLMs, trained on vast yet generic datasets, possesses an inadequate representation of the RS domain, resulting in shallow and insufficient understanding of RS imagery. Their incidental and unsystematic exposure has resulted in a limited ability to capture the rich semantic details and domain-specific knowledge required for effective RS image analysis. Therefore, the development of dedicated high-quality RS image-text datasets, coupled with specialized model architectures, plays a pivotal role in advancing the field, paving the way for more sophisticated and impactful applications.

\subsection{Remote Sensing Image-Text Paired Datasets}
\begin{figure*}
\centering
\includegraphics[width=\textwidth]{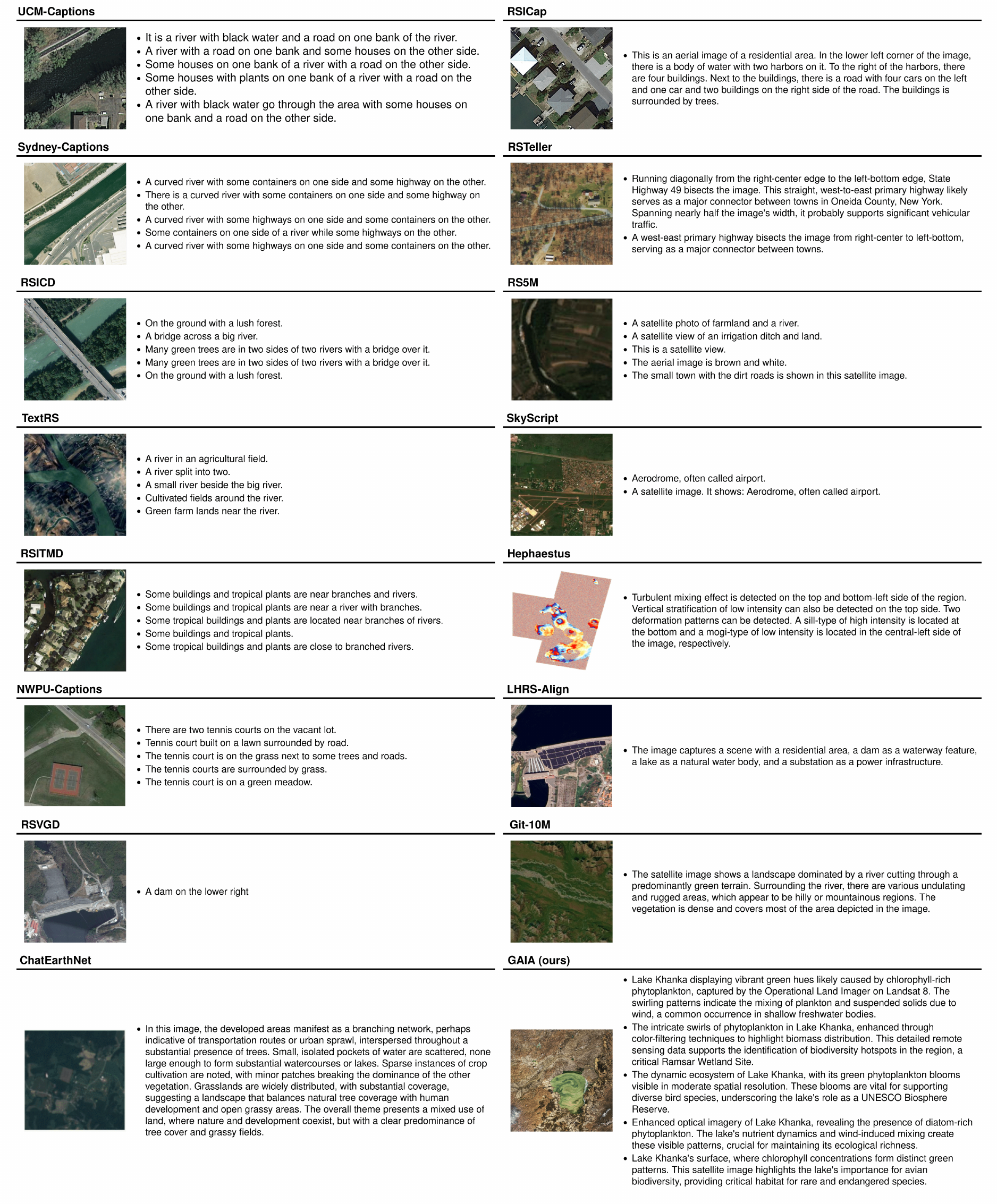}
\caption{A qualitative comparison of remote sensing (RS) image-text datasets, highlighting differences in caption length, number of captions per RS image, level of detail, and domain specificity. Notably, our GAIA dataset (bottom right) features interpretative, context-aware, and RS domain-specific captions, which differs from the predominantly object-level descriptions of datasets like UCM-Captions and RSICD.}
\label{fig:vlm_datasets}
\end{figure*}

The increasing availability of RS images, coupled with the remarkable advancements in DL, have generated unprecedented opportunities for automated satellite scene understanding, object detection, and change analysis at a global scale. However, the success of these data-intensive DL models hinges critically on the existence of large-scale, high-quality training datasets. While the computer vision community has witnessed the creation of massive image-text paired datasets like LAION-5B, the RS domain lags behind, primarily due to the inherent complexities and unique challenges associated with annotating RS imagery. Creating such datasets for RS applications (see Table~\ref{tab:datasets} and Fig.~\ref{fig:vlm_datasets}) is a formidable task, requiring domain expertise, meticulous manual effort, and substantial resources to accurately describe the complex spatial relationships, diverse land cover types, and varying scales present in aerial and satellite imagery.

Early efforts in associating RS imagery with textual descriptions were pioneered by~\cite{qu2016deep} introducing two novel datasets towards semantic understanding of high-resolution RS images: the UCM-Captions dataset and the Sydney-Captions dataset. The UCM-Captions dataset, derived from the UC Merced Land Use dataset~\cite{yang2010bag}, contains 2,100 aerial images with spatial resolution of 0.3m across 21 land-use classes, sourced from the USGS National Map Urban Area Imagery. The Sydney-Captions dataset features 613 images with spatial resolution of 0.5m across 7 land-use classes, sourced from a Google Earth. Both datasets contain five manually crafted captions for each image, describing the scene's semantic content, including objects, attributes, and their relationships. However, while the captions for each image vary in phrasing, descriptions for images within the same class show limited diversity, often focusing on similar aspects and sentence structures, revealing a limitation in capturing the diverse perspectives of human observers.

Building upon the aforementioned foundational datasets, the RSICD dataset~\cite{lu2017exploring} was introduced, significantly expanding the scale and scope of RS image-text paired datasets available at the time. RSICD contains 10,921 aerial images featuring varying spatial resolutions, collected from multiple sources including Google Earth, Baidu Map, MapABC, and Tianditu, with the objective of achieving high intra-class variance and low inter-class similarity. The images were initially annotated with 24,333 descriptive sentences, with captions per image varying from 1 to 5 and subsequently extended them to 54,605 sentences by duplicating randomly the existing sentences when there are not five different sentences to describe the same image. While RSICD represented a significant advancement in terms of size and diversity compared to UCM-Captions and Sydney-Captions, it still relied on manual annotation, which limited its scalability and required substantial human effort.

To address the need for fine-grained understanding of RS imagery, the RSITMD (Remote Sensing Image-Text Match) dataset~\cite{yuan2022exploring} was specifically designed for cross-modal RS image retrieval. RSITMD comprises of 4,743 images across 32 categories, sourced from the RSICD dataset and Google Earth. Uniquely, RSITMD emphasizes fine-grained relationships between image attributes. Multiple annotators provided five descriptive captions per image, following guidelines to document these fine-grained relationships, alongside one to five supplementary object-level keywords. Authors demonstrated that RSITMD's captions are more diverse than previous datasets, with a diversity score of 4.60, significantly higher than Sydney (1.83), UCM (0.97), and RSICD (1.67), capturing nuanced relationships between objects within images. 

Further expanding the scale and diversity of RS image captioning datasets, the NWPU-Captions dataset~\cite{cheng2022nwpu}, became the largest dataset for this task at the time. It contains 31,500 images with spatial resolution ranging from 30 to 0.2 meters, covering 45 classes based on the NWPU-RESISC45 dataset~\cite{cheng2017remote}. Each image was manually annotated with five descriptive captions by experienced RS experts, summing up to 157,500 image-caption pairs.

The TextRS dataset~\cite{abdullah2020textrs}, focused on text-to-image matching tasks within the RS domain. TextRS contains approximately 2,144 images, each associated with five descriptive sentences summing up to 10,720 image-caption pairs. The images were randomly selected from four well-established scene classification datasets: AID~\cite{xia2017aid}, UC Merced~\cite{yang2010bag}, PatternNet~\cite{zhou2018patternnet}, and NWPU~\cite{cheng2022nwpu}, covering in total 134 overlapping categories. The annotation process involved five individuals generating descriptive sentences for each image, following specific guidelines to ensure diversity, accuracy, and relevance. While TextRS addressed limitations of previous datasets by incorporating diverse imagery, its reliance on manual annotation highlighted the ongoing need for automated or semi-automated approaches.

The RSVGD dataset~\cite{zhan2023rsvg}, attempted to address the annotation bottleneck by proposing an automatic image-query generation method, although it still required manual verification to ensure correctness. The dataset was sampled from the DIOR~\cite{li2020object} dataset, originally designed for object detection, featuring 20 object categories. Authors annotation pipeline comprised of four steps: (1) box sampling, (2) attribute extraction, (3) expression generation, and 4) manual verification. The constructed dataset contains 17,402 RS images with spatial resolution ranging from 0.5m to 30m and 38,320 image-caption pairs, where each object instance in the RS image corresponds to a unique caption. The average caption length is 7.47 words and the vocabulary size is 100 words.

Hephaestus~\cite{Bountos_2022_CVPR}, a dataset designed for InSAR image analysis, offers a rich collection of manually annotated Sentinel-1 interferograms specifically designed for diverse computer vision tasks, including image captioning. The dataset has been manually annotated by a team of InSAR experts and contains 19,919 individual interferograms over 44 volcanoes across the globe, with each interferogram paired with a single, yet detailed caption. These captions provide comprehensive textual descriptions of each interferogram, including observed ground deformation, atmospheric contributions, and overall image quality. The InSAR images were acquired from the Comet-LiCS~\cite{lazecky2020licsar} portal, an automated system that processes Sentinel-1 data to generate interferograms.

Marking a significant leap towards fully automated image-caption dataset creation, the RSTeller dataset~\cite{ge2024rsteller}, provided approximately 1.2 million image patches, each paired with multiple captions, resulting in more than 2.5 million RS image-text pairs, with a ground sampling distance of 0.6 meters. The imagery was sourced exclusively from the National Agriculture Imagery Program (NAIP) via the Google Earth Engine platform. The temporal coverage spans from August 1, 2021, to November 26, 2022, with all imagery captured via aircraft over the United States. RSTeller employed an automated workflow consisting of four main processes: (1) raw data acquisition from Google Earth Engine and OpenStreetMap (OSM), (2) raw caption generation using the Mixtral-7B large language model, (3) caption augmentation for linguistic diversity, and (4) quality control. The caption lengths varied from 5 to 192 tokens, with a median length of 48 tokens and an average of 54.2 tokens per caption. While RSTeller demonstrated the feasibility of large-scale automated dataset creation, its focus on NAIP imagery limited its diversity in terms of geographic coverage and image sources.

The RS5M~\cite{zhang2024rs5m} dataset, represented a major advancement in scale, providing a staggering 5 million image-text pairs. Its creation employed a two-stage methodology. First, a comprehensive filtering process was applied to 11 publicly available image-text datasets from the broader computer vision domain. This filtering stage, resulting in the PUB11 subset, involved text-based keyword filtering, de-duplication, and a dual-phase filtering process leveraging a pre-trained VLM to exclude images unrelated to RS. Second, the RS3 subset was constructed from existing class-labeled RS datasets, namely BigEarthNet-v1~\cite{sumbul2019bigearthnet, sumbul2021bigearthnet}, FMoW~\cite{christie2018functional}, and MillionAID~\cite{long2021creating}. Descriptive captions were generated for these datasets using a pre-trained BLIP2~\cite{li2023blip} model, and their quality was enhanced using a ranking system based on CLIP models and a novel rotational-invariance criterion. Finally, relevant meta-information, such as class labels, UTM coordinates, and UTC timestamps, were integrated into the captions, where available. The resulting RS5M dataset boasted a diverse range of image types, including both satellite and aerial perspectives, paired with descriptive English language captions often further enhanced with geographic, temporal, and scene-specific meta-information.

SkyScript~\cite{wang2024skyscript}, addressed the need for semantic richness by comprising 2.6 million image-text pairs with an impressive 29 thousand unique tags derived from OpenStreetMap (OSM). These tags detailed object categories, subcategories, and fine-grained attributes, providing a level of detail not found in previous datasets. SkyScript's imagery was sourced from Google Earth Engine (GEE) collections, with spatial resolutions spanning 0.1m to 30m per pixel. The dataset's global coverage favored regions with abundant high-resolution imagery and comprehensive OSM annotations, such as the U.S. and Europe. An automated annotation pipeline connected images to OSM data, employing a two-stage object selection for diverse representation, image selection based on object-specific criteria, and a two-stage tag classification using CLIP embeddings to ensure visual relevance and appropriate ground sampling distance. Rule-based caption generation then synthesized OSM tags into descriptive captions, while a final filtering step, based on CLIP-derived image-text similarity, refined the dataset's quality. While SkyScript represented a significant advancement in terms of semantic richness and automated annotation, its coverage bias towards regions with rich OSM data presented a limitation.

LuojiaHOG~\cite{zhao2024luojiahog}, focused on RS image-text retrieval, offering a dataset of 94,856 images sourced from Google Earth, covering diverse regions globally. A key feature of LuojiaHOG was its hierarchical classification system aligned with Open Geospatial Consortium (OGC) standards, featuring 7 first-level, 21 second-level, and 131 third-level categories, allowing for extensibility and compatibility with various data under different requirements. The dataset employed a hybrid annotation approach, combining manual and automatic processes. Manual annotation involved expert correction of OpenStreetMap (OSM) labels and the creation of detailed image descriptions following specific guidelines. Automatic annotation utilized Minigpt4~\cite{zhu2023minigpt} with prompt engineering to generate a large volume of image descriptions, with quality control measures implemented. Statistical analysis revealed an average caption length of 123.56 words and 6.95 sentences per caption. While LuojiaHOG's hierarchical structure and detailed annotations are valuable, the hybrid annotation approach still involved significant manual effort and limited its scalability.

RSICap and RSIEval~\cite{HU2025272}, two interconnected datasets were designed for training and evaluating vision-language models in RS. RSICap consists of 2,585 high-quality, human-annotated captions. The dataset included imagery derived from the DOTA~\cite{xia2018dota} object detection dataset, featuring different satellite and aerial sensors such as GF-2, JL-1 and Google Earth, encompassing both color and panchromatic images with a wide range of spatial resolutions. The annotation process followed specific guidelines, including describing image attributes, object attributes, and overall scene, as well as including reasonable speculations based on visual content. RSICap captions have an average length of 60 words. RSIEval, on the other hand, was designed for benchmarking VLMs on RS image captioning (RSIC) and visual question answering (RSVQA) tasks. It contains 100 high-quality image-caption pairs (following RSICap annotation guidelines) and 936 diverse image-question-answer triplets, averaging nine questions per image. The questions covered four categories: object-related, image-related, scene-related, and reasoning-related. While RSICap and RSIEval provided valuable resources for training and evaluation, their reliance on manual annotation limited their scalability.

ChatEarthNet~\cite{yuan2024chatearthnet}, utilized ChatGPT-3.5 and ChatGPT-4V to generate 163,488 and 10,000 image-text pairs from Sentinel-2 imagery, respectively. The imagery featured global coverage and diverse temporal distribution across all continents except Antarctica, including nine spectral bands. Semantic information was derived from the ESA WorldCover project, which provides land cover maps at 10-meter resolution with 11 classes. ChatEarthNet features detailed natural language descriptions for each image, guided by prompts that incorporated land cover proportions and spatial distributions, offering a mean of 90 words per caption for ChatGPT-4V and 155 for ChatGPT-3.5. While ChatEarthNet demonstrated the potential of large language models for automated caption generation, its reliance on specific data sources (Sentinel-2 and ESA WorldCover) still presented limitations in terms of diversity.

LHRS-Align~\cite{muhtar2024lhrs} is a large-scale dataset of 1.15 million RS image-caption pairs. Images, sourced from Google Earth at 1m resolution, were paired with geographic features from OpenStreetMap (OSM). The dataset's annotation pipeline involved geo-aligning images with OSM features, pruning irrelevant attributes, and generating captions using Vicuna-v1.5-13B. While the dataset offers a high degree of automation and addresses diverse tasks, the reliance on the effectiveness of a single language model highlights ongoing challenges.

Finally, Git-10M~\cite{10988859}, a large-scale RS image-text paired dataset contains 10 million pairs, sourced by 30\% from public datasets (Million-AID~\cite{long2021creating}, GeoPile~\cite{mendieta2023towards}, SSL4EO-S12~\cite{wang2023ssl4eo}, SkySript~\cite{wang2024skyscript}) and 70\% from Google Earth imagery. The dataset spans diverse geographical regions globally and includes various spatial resolutions ranging from 0.5m to 128m/pixel and geospatial metadata. They authors employ an automated annotation pipeline using GPT-4o, with prompts incorporating included geospatial metadata to generate semantically accurate descriptions.

In conclusion, the evolution of image-text paired datasets in RS has been marked by a progression towards larger scales, increased diversity, and greater automation. Early datasets like UCM-Captions and Sydney-Captions laid the groundwork for RS image captioning but suffered from limited diversity and relied heavily on manual annotation. Subsequent datasets like RSICD, RSITMD, and NWPU-Captions expanded the scale and introduced more fine-grained descriptions, but manual annotation remained a significant bottleneck. The emergence of datasets like RSVGD, RSTeller, RS5M, SkyScript, LuojiaHOG and Git-10M represented a shift towards semi-automated and fully automated annotation approaches, leveraging techniques like language models, rule-based systems, and filtering based on visual-language models. However, challenges persist in terms of scalability, the labor and expertise required for annotation (even in semi-automated approaches), and the need for datasets that are both diverse and specific to the unique characteristics of RS imagery. Addressing these challenges will require continued research into novel annotation methodologies, the development of more sophisticated vision-language models tailored for RS, and the creation of comprehensive benchmarks that capture the full spectrum of RS tasks.

\begin{table*}[ht]
\centering
\caption{Summary of Remote Sensing Image-Text Paired Datasets}
\label{tab:datasets}
\resizebox{0.95\textwidth}{!}{%
\begin{tabular}{l|c|c|c|>{\centering\arraybackslash}p{1.5cm}|>{\centering\arraybackslash}p{2.0cm}|>{\centering\arraybackslash}p{1.5cm}|>{\centering\arraybackslash}p{1.6cm}|>{\centering\arraybackslash}c|>{\centering\arraybackslash}c|>{\centering\arraybackslash}p{2.8cm}|>{\centering\arraybackslash}c}
\toprule
\multicolumn{1}{c|}{\multirow{2}{*}{\textbf{Dataset}}} &
\multicolumn{1}{c|}{\multirow{2}{*}{\textbf{Year}}} &
\multicolumn{1}{c|}{\multirow{2}{*}{\textbf{Images}}} &
\multicolumn{1}{c|}{\multirow{2}{*}{\begin{tabular}[c]{@{}c@{}}\bfseries Image-Text\\\bfseries Pairs\end{tabular}}} &
\multicolumn{1}{c|}{\multirow{2}{*}{\begin{tabular}[c]{@{}c@{}}\bfseries Spatial\\\bfseries Resolution\\\bfseries (m/pixel)\end{tabular}}} & % Col 5
\multicolumn{1}{c|}{\multirow{2}{*}{\begin{tabular}[c]{@{}c@{}}\bfseries Spatial\\\bfseries Coverage\end{tabular}}} & % Col 6
\multicolumn{1}{c|}{\multirow{2}{*}{\begin{tabular}[c]{@{}c@{}}\bfseries Temporal\\\bfseries Coverage\end{tabular}}} & % Col 7
\multicolumn{1}{c|}{\multirow{2}{*}{\begin{tabular}[c]{@{}c@{}}\bfseries Image\\\bfseries Modalities\end{tabular}}} & % Col 8
\multicolumn{2}{c|}{\textbf{Observation Type}} & % Cols 9-10
\multicolumn{1}{c|}{\multirow{2}{*}{\textbf{Data Source(s)}}} & % Col 11
\multicolumn{1}{c}{\multirow{2}{*}{\begin{tabular}[c]{@{}c@{}}\bfseries Annotation\\\bfseries Method\end{tabular}}} \\ % Col 12
\cmidrule(lr){9-10}
& & & & & & & & % 8 ampersands for columns 1-8
\multicolumn{1}{c|}{\begin{tabular}[c]{@{}c@{}}\bfseries Standard\end{tabular}} & % Content for Col 9
\multicolumn{1}{c|}{\begin{tabular}[c]{@{}c@{}}\bfseries Event-Triggered\end{tabular}} \\ % Content for Col 10. No & after this, as Cols 11-12 are spanned from above.
\midrule
UCM-Captions~\cite{qu2016deep} & 2016 & 2,100 & 10,500 & 0.3 & USA & - & Optical (RGB) & \checkmark & $\times$ & USGS National Map Urban Area Imagery & Manual \\
\midrule
Sydney-Captions~\cite{qu2016deep} & 2016 & 613 & 3,065 & 0.5 & Sydney, Australia & - & Optical (RGB) & \checkmark & $\times$ & Google Earth & Manual \\
\midrule
RSICD~\cite{lu2017exploring} & 2017 & 10,921 & 54,605 & Variable & - & - & Optical (RGB) & \checkmark & $\times$ & Google Earth, Baidu Map, MapABC, Tianditu & Manual \\
\midrule
TextRS~\cite{abdullah2020textrs} & 2020 & 2,144 & 10,720 & Variable & - & - & Optical (RGB) & \checkmark & $\times$ & AID, UC Merced, PatternNet, NWPU & Manual \\
\midrule
RSITMD~\cite{yuan2022exploring} & 2022 & 4,743 & 23,715 & Variable & - & - & Optical (RGB) & \checkmark & $\times$ & RSICD, Google Earth & Manual \\
\midrule
NWPU-Captions~\cite{cheng2022nwpu} & 2022 & 31,500 & 157,500 & 30 -- 0.2 & - & - & Optical (RGB) & \checkmark & $\times$ & NWPU-RESISC45 & Manual \\
\midrule
RSVGD~\cite{zhan2023rsvg} & 2023 & 17,402 & 38,320 & 0.5 -- 30 & - & - & Optical (RGB) & \checkmark & $\times$ & DIOR & Hybrid \\
\midrule
Hephaestus~\cite{Bountos_2022_CVPR} & 2023 & 19,919 & 19,919 & 100 & 44 most active volcanoes globally & 2014-2021 & Radar (InSAR) & \checkmark & \checkmark & Comet-LiCS & Manual \\
\midrule
RSICap~\cite{HU2025272} & 2023 & 2,585 & 2,585 & Variable & - & - & Optical (RGB) & \checkmark & $\times$ & DOTA, GF-2, JL-1, Google Earth & Manual \\
\midrule
RSTeller~\cite{ge2024rsteller} & 2024 & 1.2M & 2.5M & 0.6 & USA & 2021-2022 & Optical (RGB) & \checkmark & $\times$ & NAIP via Google Earth Engine & Automatic \\
\midrule
RS5M~\cite{zhang2024rs5m} & 2024 & 5M & 5M & Variable & Global & - & Optical (RGB) & \checkmark & $\times$ & \textbf{PUB11 filtered datasets}: LAION-400M, YFCC100M, CC3M, CC12M, RedCaps, WIT, TextCaps, VizWiz, COCO, SBU, Conceptual Captions\newline\newline \textbf{RS3 datasets}: BigEarthNet-v1, FMoW, MillionAID & Automatic \\
\midrule
SkyScript~\cite{wang2024skyscript} & 2024 & 2.6M & 2.6M & 0.1 -- 30 & Global & - & Optical (RGB) & \checkmark & $\times$ & Google Earth Engine & Automatic \\
\midrule
LuojiaHOG~\cite{zhao2024luojiahog} & 2024 & 94,856 & 94,856 & Variable & Global & - & Optical (RGB) & \checkmark & $\times$ & Google Earth & Hybrid \\
\midrule
ChatEarthNet~\cite{yuan2024chatearthnet} & 2024 & 173,488 & 173,488 & 10 & Global & 2020 & Optical (Multi-spectral) & \checkmark & $\times$ & Sentinel-2, ESA WorldCover & Automatic \\
\midrule
LHRS-Align~\cite{muhtar2024lhrs} & 2024 & 1.15M & 1.15M & 1 & Global & - & Optical (RGB) & \checkmark & $\times$ & Google Earth, OSM & Automatic \\
\midrule
Git-10M~\cite{10988859} & 2025 & 10M & 10M & 0.5 -- 128 & Global & - & Optical (RGB) & \checkmark & $\times$ & \textbf{30\% public datasets} (Million-AID, GeoPile, SSL4EO-S12, SkyScript), \textbf{70\% Google Earth} & Automatic \\
\midrule
\midrule
GAIA (ours) & 2025 & 40,201 & 201,005 & Variable & Global & 1998-2024 & Multi-modal Visual Composites & \checkmark & \checkmark & Web-scraped articles (images \& text) from several RS-related websites & Automatic \\
\bottomrule
\end{tabular}
}
\end{table*}

\subsection{The imperative need for better annotations}
The performance and efficacy of VLMs are critically dependent on two key factors: the sophistication of their architectural design and the quality, diversity, and scale of the training data. OpenAI’s CLIP~\cite{radford2021learning}, for instance, revolutionized the field with its unprecedented ability to bridge vision and language modalities, setting a new benchmark for multi-modal learning. However, while the model weights were made publicly available, the specific dataset used for training remains undisclosed, leaving a gap in the understanding of the data-driven foundations of its success. LAION-AI with OpenCLIP~\cite{ilharco_gabriel_2021_5143773}, the open-source implementation of OpenAI's CLIP~\cite{radford2021learning}, has demonstrated impressive results. They managed to replicate OpenAI's proprietary pre-training dataset~\cite{schuhmann2022laion} and subsequently trained and published several models~\cite{cherti2023reproducible}, using various architectures, on a variety of data sources and compute budgets, ranging from small to large-scale experiments. Recent advancements in the context of CLIP pre-training have showcased remarkable achievements along the axes of pre-train data refinement~\cite{gadre2024datacomp, fang2023data, xu2023demystifying}, model architecture~\cite{sun2023eva} and computational efficiency~\cite{li2023inverse, li2023clipa, zhai2023sigmoid}, leading to substantial improvements and eventually establishing new standards within the era of pre-trained CLIP models.

Despite the appeal of web-scraped image-text data as a cost-effective resource for encoding general knowledge, their inherent noise and the finite nature of the internet archive itself pose significant limitations. Recognizing that data quality often supersedes sheer quantity~\cite{birhane2024into}, recent works have been focusing on refining these datasets through advanced filtering and annotation improvement techniques. This pursuit has been significantly accelerated by the advent of foundation models, which have enabled a new wave of innovative approaches to filter and enhance existing datasets.
 
Early endeavors concentrated on the expansion of datasets through the incorporation of synthetic captions. A pioneering effort in this direction was the creation of LAION-COCO~\cite{laioncoco}, a dataset encompassing 600 million synthetic captions derived from the LAION2B-en dataset. This work underscored the viability of employing synthetic data to augment the breadth and heterogeneity of image-text datasets.

Subsequent investigations have explored a variety of techniques aimed at refining and augmenting existing datasets. One prominent approach involves improving the caliber of captions through the utilization of sophisticated language models. Leveraging language rewrites has been shown to improve the quality of captions~\cite{fan2024improving}. In one such approach, a large language model (GPT-3) was used to rewrite captions from the Conceptual Captions and SBU Captions datasets to enhance their descriptiveness and accuracy. This method was shown to improve the performance of CLIP models on downstream tasks, demonstrating the importance of high-quality captions for contrastive language-image pre-training.

Concurrently, related work has focused on refining captions in multi-modal datasets to improve model training~\cite{nguyen2024improving}. In this approach, a state-of-the-art vision-language model (BLIP-2) was used to generate synthetic captions. Incorporating these new captions during training was found to lead to significant performance gains in downstream tasks, further highlighting the importance of caption quality.

Another notable contribution in this area is the ShareGPT4V system~\cite{chen2025sharegpt4v}, which aimed to enhance the performance of large multi-modal models through high-quality image captions. This approach involved fine-tuning a large language model on a carefully curated dataset of image-caption pairs, demonstrating the benefits of meticulously curated datasets for improving the performance of large-scale models. The study empirically showed that replacing just 3.5\% of standard supervised fine-tuning data with their high-quality captions led to significant performance improvements in existing models like LLaVA-1.5 and Qwen-VL-Chat.

In a paradigm shift, CapsFusion~\cite{yu2024capsfusion} demonstrated a novel approach that rethinks image-text data at scale. CapsFusion leverages an ensemble of captioning models, including BLIP, BLIP-2, and GIT, to generate a diverse set of captions for each image. These captions are then fused using a large language model, specifically Vicuna-7B, to create a more comprehensive description. This method addresses the limitations of single-model captioning and demonstrates the advantages of leveraging multiple models to enhance caption quality.

The potential of alt-text as a valuable source of image descriptions, has also been investigated~\cite{xu2024altogether}, introducing a method for re-aligning alt-text with images and thereby leveraging this often underutilized information source. They employed a multi-stage training process involving human annotation and a specifically designed captioning model to enhance the alignment between alt-text and image content. This method was shown to improve image captioning performance, highlighting the value of alt-text in providing rich visual descriptions.

The application of synthetic data was further explored in specialized domains, such as medical vision-language pre-training~\cite{liu2024can}. They investigated the feasibility of training effective medical vision-language models using purely synthetic data. They created a synthetic dataset of chest X-ray images and reports using generative models and demonstrated that models trained on this synthetic data could achieve comparable performance to those trained on real data, thus addressing data scarcity and privacy concerns inherent in medical datasets.

FuseCap~\cite{rotstein2024fusecap} introduced a system that leverages large language models to generate enriched, fused image captions. FuseCap employs frozen vision encoders, including an object detector, an attribute recognizer, and an OCR model, to extract detailed visual information from images. These outputs are then fused with original captions using a fine-tuned Flan-T5 model, resulting in comprehensive image descriptions. They curated a dataset of 12 million image-enriched caption pairs and demonstrated that training a BLIP-based captioning model on this data led to improved performance on downstream tasks.

Advancing the use of synthetic data, a method that boosts visual-language models by utilizing synthetic captions and image embeddings was introduced~\cite{sharifzadeh2024synth}. This approach employs a text-to-image model (MUSE) to generate synthetic image embeddings from captions generated by a large language model (LLaMA-3). They demonstrated that fine-tuning a VLM on this synthetic data achieves comparable performance to models trained on real data, highlighting the efficacy of combining different types of synthetic data.

Finally, a large-scale experiment to investigate the capabilities of state-of-the-art language models in generating high-quality captions has been conducted~\cite{li2024if}. A LLaMA-3-powered LLaVA model was employed to recaption 1.3 billion web images from the DataComp-1B dataset. Their results demonstrated significant improvements in the quality of the generated captions compared to the original web-crawled captions, highlighting the potential of advanced language models in large-scale data augmentation.

Despite these advancements, specialized fields such as medical imaging and remote sensing continue to face challenges due to the complex, domain-specific knowledge required for data curation, and the significantly lower tolerance for errors. In these high-stake domains, inaccuracies can have profound consequences. To this end, the medical domain has addressed this challenge by leveraging biomedical image-caption pairs, meticulously collected and filtered from various sources including open-access research publications \cite{pelka2018radiology, subramanian2020medicat, lin2023pmc, liu2023qilin}, medical reports \cite{johnson2019mimic, bustos2020padchest, li2021ffa}, and even social media platforms \cite{huang2023visual, ikezogwo2023quilt}, as a trustworthy source of paired image-text data. In contrast, the RS domain has been lately limited to publicly available annotation like land cover maps (e.g. ESA's WorldCover) in conjunction with open-access RS imagery and large language models (LLMs) to generate grounded image-text pairs.

The research trajectory aimed at refining annotated datasets has dramatically progressed, yet challenges persist within specialized domains, particularly given their strict requirements for accuracy. Given the powerful capabilities of current foundation models, it is imperative for the RS community to explore and harness more effective priors. Drawing inspiration from the medical domain's approach, the focus should shift towards leveraging these priors, and with the aid of the current ever-evolving foundation models, restructuring this information to develop more robust and adaptable in-domain vision-language models.

\section{Description of GAIA}
\subsection{Dataset Construction}
\begin{figure}[h]
\centering
\includegraphics[width=\columnwidth]{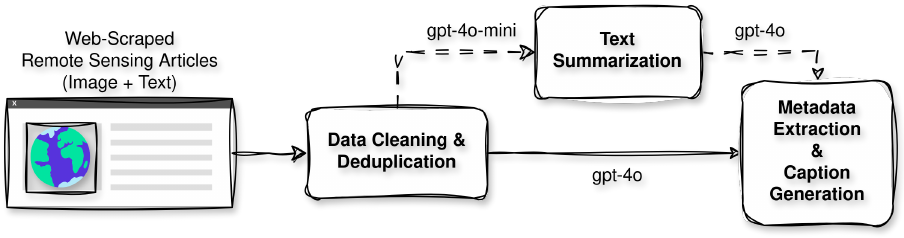}
\caption{Overview of GAIA data acquisition and annotation pipeline. This figure illustrates the process of building the GAIA dataset, from initial data acquisition to the generation of enriched metadata and captions. The pipeline begins with web-scraped RS articles (image + text) as the foundational data. This raw data undergoes a rigorous data cleaning and de-duplication process, which also involves text summarization using GPT-4o-mini in cases of large articles. Subsequently, GPT-4o is employed to extract metadata from the cleaned data and generate more descriptive captions, resulting in the comprehensive GAIA dataset.}
\label{fig:annotation_pipeline}
\end{figure}

To address the scarcity of high-quality, domain-specific image-text paired datasets in RS, we introduce GAIA, a novel dataset constructed through an automated, yet robust annotation pipeline depicted in Fig.~\ref{fig:annotation_pipeline} to advance RS image understanding and analysis. GAIA's data collection pipeline involves targeted web-scraping of image-text pairs from reputable RS-related sources, followed by rigorous data cleaning and de-duplication, to ensure data integrity. For the data annotation part, we leverage ChatGPT and more specifically GPT-4o (omni). Through carefully crafted prompts, we instruct the model to extract domain-specific metadata and ultimately generate five distinct and diverse in-domain captions for each RS image, ensuring that each caption accurately interprets the visual data and provides semantically rich descriptions of the RS imagery. In contrast to existing RS image-text paired datasets that feature arbitrary RS imagery, GAIA provides a curated collection of semantically rich data. GAIA distinguishes itself through its global coverage, encompassing both anthropogenic and natural occurring events. This targeted approach enables the development of domain-specific VLMs capable of learning abstract representations that effectively encode the complex features and dynamic processes shaping our planet.

\begin{figure*}[h]
\centering
\includegraphics[width=\textwidth]{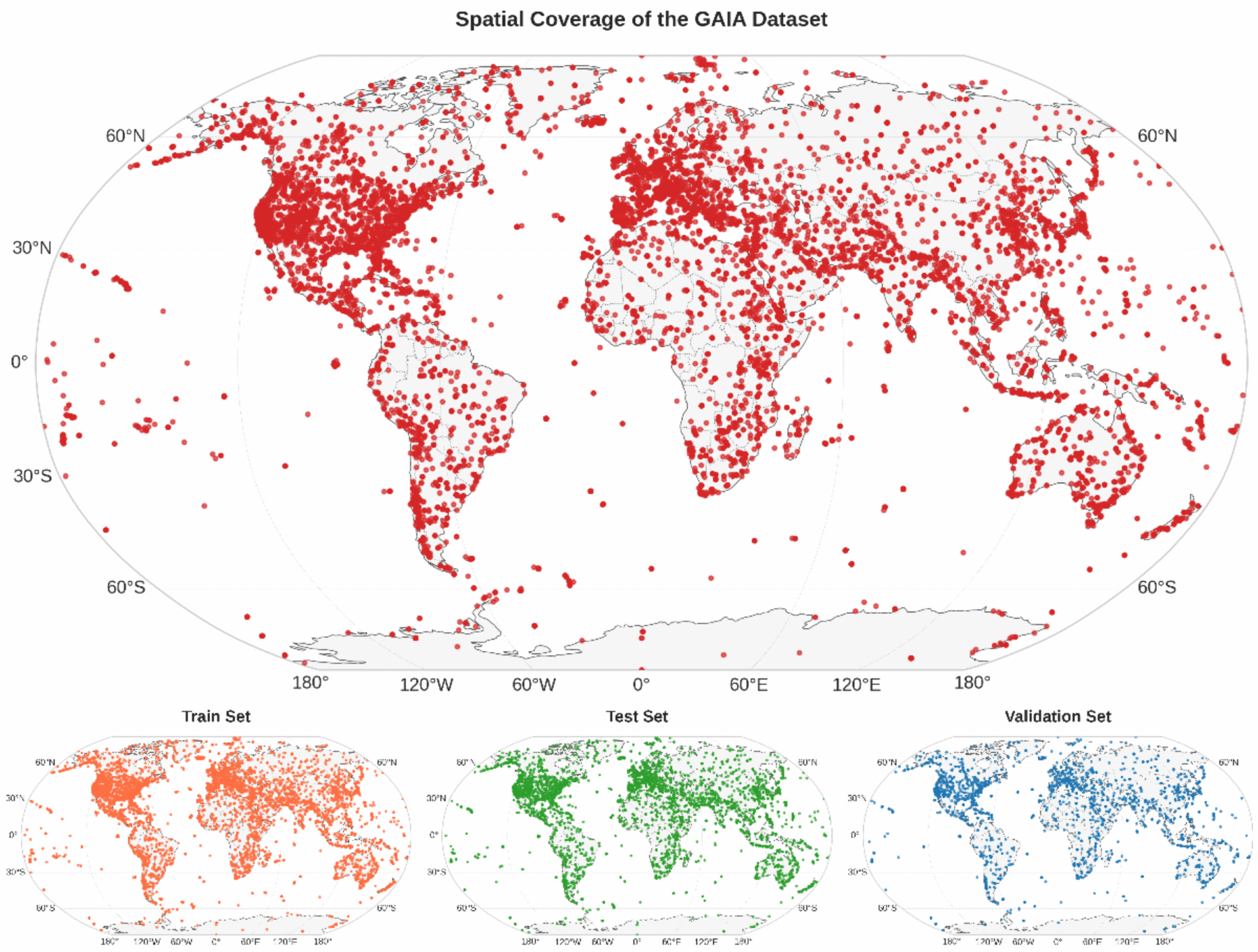}
\caption{Spatial coverage and distribution of the full GAIA dataset, in conjunction with the train, test, and validation sets. The main figure (top) illustrates the global spatial distribution of samples in the GAIA dataset. Each red dot represents a location associated with an image-text pair. The dataset exhibits broad coverage across various regions, with higher concentrations observed over North America, Europe, and parts of Asia and South America, including the often neglected Antarctic region. The three smaller maps (bottom) display the spatial distribution for the train set (orange), test set (green), and validation set (blue), respectively. This visualization demonstrates that the GAIA dataset provides a geographically diverse representation of Earth's surface, which is crucial for training and evaluating robust RS models. The train, test, and validation sets maintain a similar spatial distribution pattern to the overall dataset, ensuring consistency across different data splits.}
\label{fig:spatial_coverage}
\end{figure*}

\subsubsection{Data Collection}
The data collection process was designed to be both comprehensive and efficient, ensuring high-quality and diverse data acquisition. We utilized custom web-scraping scripts to extract images and text from a curated list of reputable online sources which focus on the dissemination of Earth Observation articles that incorporate and analyze RS imagery. These sources were selected based on their data relevance, quality and diversity of content and are mainly comprised of online image repositories maintained by leading Earth Observation organizations. Due to the fact that most of these sources contain highly unstructured information, our web scraping efforts concentrated on extracting images, along with rich metadata and detailed descriptions, creating a solid foundation for subsequent annotation. We strictly adhered to the terms of service and licensing agreements of the respective sources. Our published dataset includes proper attribution to data providers and article authors within the metadata accompanying each sample. The sources utilized in this study, along with the respective counts of web-scraped articles, are summarized in Table~\ref{tab:web_scrape_counts}, while Table~\ref{tab:metadata} provides an overview of the metadata fields collected for each sample, when available.

\begin{table*}[h]
\centering
\renewcommand{\arraystretch}{1.3}
\caption{Article Counts Before and After Filtering by Source}
\label{tab:web_scrape_counts}
\begin{tabular}{llrrr}
\hline
\textbf{Source} & \textbf{Domain(s)} & \textbf{Original} & \textbf{Filtered} & \textbf{\% Retained} \\
\hline
\multirow{2}{*}{NASA} & visibleearth.nasa.gov & \multirow{2}{*}{54,577} & \multirow{2}{*}{35,468} & \multirow{2}{*}{65.0} \\
                      & earthobservatory.nasa.gov & & & \\
\multirow{2}{*}{ESA}  & esa.int & \multirow{2}{*}{5,932} & \multirow{2}{*}{3,487} & \multirow{2}{*}{58.8} \\
                      & copernicus.eu & & & \\ 
NOAA    & nesdis.noaa.gov & 651 & 438 & 67.3 \\ 
AIRBUS  & space-solutions.airbus.com & 413 & 409 & 99.0 \\ 
Planet  & planet.com & 612 & 399 & 65.2 \\ \hline
Total   & & 62,185 & 40,201 & 66.0 \\ \hline
\end{tabular}
\end{table*}

\subsubsection{Data Cleaning and De-duplication} \label{sec:data_cleaning}
The raw web-scraped dataset underwent a data cleaning and de-duplication phase. Textual data was cleaned by removing HTML tags and URLs. To address instances of empty alt-text HTML tags, the corresponding article's title was stored as a substitute of the original image caption. Image de-duplication was performed in three distinct phases. Initially, duplicate images were identified based on their source URLs. Subsequently, we utilized fastdup~\footnote{\url{https://github.com/visual-layer/fastdup}}, an open-source tool to identify more duplicates, outliers, and corrupted images. In cases of duplication, dataset entries were merged, including their metadata, while outliers and corrupted images were eliminated from the dataset. Ultimately, we implemented a multi-stage filtering protocol to further enhance dataset quality by identifying and removing remaining outliers and low-quality samples. For this, we leveraged a state-of-the-art vision-language model, PE-Core-G14-448 \cite{bolya2025perception}, to score each image based on its semantic similarity to a set of text prompts (e.g., 'a satellite image', 'an earth observation image', etc.). Our filtering process proceeded in three stages. First, ~30 images with extremely low similarity scores were flagged; manual inspection confirmed these were remaining outliers, and they were subsequently removed. Second, to address images of insufficient quality, we manually evaluated the next 4,000 lowest-scoring samples, representing approximately 10\% of the dataset. This rigorous review led to the removal of ~800 images that, while relevant, failed to meet our quality standards. Criteria for removal included poor visual fidelity (e.g., low-resolution or blurry images), the presence of rendered content, and scientific visualizations that lacked essential interpretative elements like legends or labels. Finally, to validate the integrity of the resulting dataset, we performed a final quality check on a random subset of 1,000 images. This final inspection confirmed the effectiveness of our data cleaning and de-duplication protocol, as no further issues were identified.

\subsubsection{Data Annotation Pipeline}
To facilitate the extraction of relevant metadata and the generation of comprehensive captions, a detailed prompt was developed to instruct ChatGPT and specifically their most capable model GPT-4o (gpt-4o-2024-08-06). The model was tasked with analyzing each RS image and its associated article's text, extracting relevant domain-specific metadata, and generating five distinct, detailed captions per image. Following OpenAI's best practices guide, we defined effective instructions for the model, such that it consistently generates content that meets our requirements and a structured output schema that ensures the model will always generate responses that adhere to our supplied JSON schema, without omitting or hallucinating required fields. The prompt employed a system of rewards and penalties to reinforce desirable and discourage undesirable caption characteristics, respectively. This structured approach promotes the extraction of domain-specific metadata and captions, enhancing consistency and facilitating downstream tasks such as dataset querying, filtering, and analysis. The systematic combination of detailed prompting and structured output contributed to the creation of a rich and informative dataset tailored for EO and RS applications. An overview of the extracted metadata fields and their descriptions is presented in Table~\ref{tab:metadata}. For excessively long articles, a text summarization step was prepended to the data annotation pipeline. This involved utilizing the far more cost-effective, yet capable for the summarization task, GPT-4o-mini (gpt-4o-mini-2024-07-18) model to generate concise summaries, thereby mitigating annotation costs for the following annotation pipeline step while preserving the quality of extracted metadata and captions. Our qualitative analysis revealed that the geographic coordinates (EPSG:4326) generated by GPT-4o, while generally proximate, lacked the desired precision. Consequently, the GPT-4o generated coordinates were discarded in favor of those obtained from the Google Geocoding API~\footnote{\url{https://developers.google.com/maps/documentation/geocoding/requests-geocoding}}, which provides accurate latitude and longitude coordinates through toponym resolution.

\subsubsection{Overlap with Existing Datasets}
A key contribution of the GAIA dataset lies in its novelty, which was evaluated by quantifying its overlap with currently available web-crawled datasets (see Section~\ref{sec:seen_rs}). Applying the consistent two-step data cleaning and de-duplication process outlined in Section~\ref{sec:data_cleaning}, we determined that GAIA introduces a substantial proportion of previously unseen images. Specifically, less than 3\% of GAIA's images are found in the compared web-crawled datasets. These overlapping samples were deliberately retained, as GAIA provides five additional captions per image, augmenting the sources' alt-text caption available in the existing datasets.

\begin{table*}[h]
\renewcommand{\arraystretch}{1.3}
\centering
\caption{Metadata Fields in the GAIA Dataset: Web-scraped and synthetically-generated. The GAIA dataset includes two types of metadata: (1) original metadata scraped from web sources during data collection, and (2) additional metadata generated by GPT-4o to enhance the dataset. The table is divided into two sections to represent each type. This comprehensive metadata approach provides rich contextual information for each data sample and facilitates detailed analysis.}
\label{tab:metadata}
\begin{tabular}{>{\ttfamily}ll}
\toprule
\multicolumn{2}{c}{\textbf{Part 1: Web-scraped Metadata}} \\
\midrule
\textbf{Metadata Field} & \textbf{Description} \\
\midrule
id & Unique identifier for each data sample. \\
image\_src & URL of the hosted image file. \\
image\_alt & Original alt-text provided for the image. \\
datetime & Date and time associated with the image. \\
location & Geographic location associated with the image. \\
coordinates & Geographic coordinates (latitude, longitude) in EPSG:4326 format. \\
topics & Relevant keywords or subject tags. \\
satellite & Satellite used to acquire the image. \\
credits &  Information on image data source, and text authorship. \\
\midrule
\multicolumn{2}{c}{\textbf{Part 2: Synthetically-generated Metadata}} \\
\midrule
\textbf{Metadata Field} & \textbf{Description} \\
\midrule
tag & Up to three domain-specific tags that describe the image. \\
location & Precise point location, preferably in 'Landmark, City, Country' format. \\
coordinates & Approximate geographical coordinates of the point location in EPSG:4326 format. \\
modality & Remote Sensing modalities or sensor types used (e.g., Optical, Thermal). \\
satellite & Name of the Remote Sensing satellite (e.g., Sentinel-2). \\
sensor & Remote Sensing sensor/instrument used (e.g., MultiSpectral Instrument (MSI)). \\
resolution & Spatial resolution of the sensor: Low, Moderate, High, or Very High. \\
\bottomrule
\end{tabular}
\end{table*}

\subsection{Exploratory Data Analysis}
\label{sec:eda}
\begin{figure*}
\centering
\includegraphics[width=\textwidth]{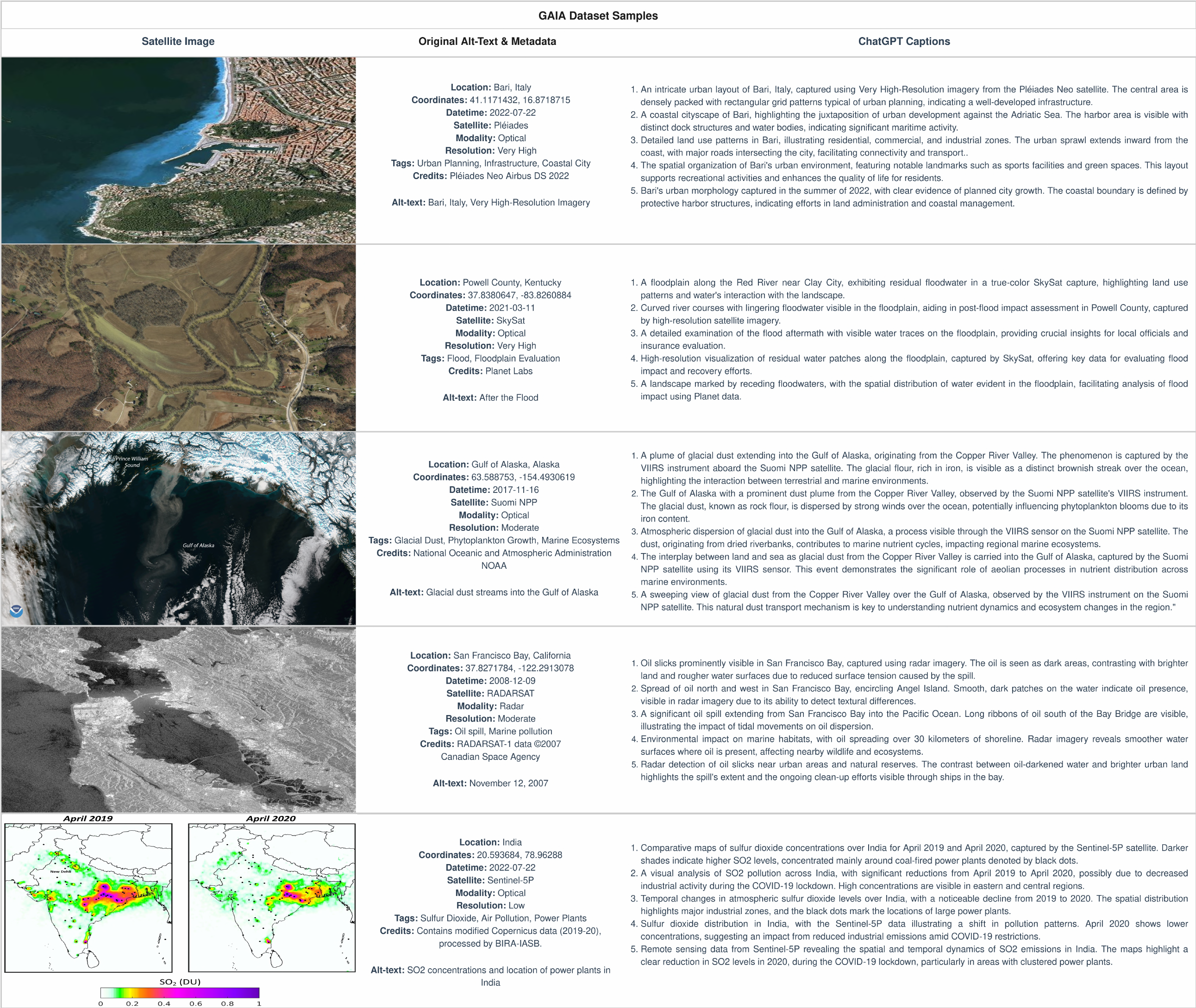}
\caption{A glimpse into the GAIA dataset, showing various satellite images alongside metadata, the original alt-text and our five descriptive synthetic captions. This diverse content highlights the dataset's heterogeneity, as well as the enhanced descriptive richness achieved through our synthetic captions.}
\label{fig:gaia_samples}
\end{figure*}

This section presents a detailed exploratory data analysis of GAIA, delving into its temporal, spatial, and spectral characteristics, as well as the semantic content of its captions. The insights gained from this analysis, along with the illustration of GAIA dataset samples in Fig.~\ref{fig:gaia_samples}, underscore the unique strengths and potential applications of the dataset for training and evaluating advanced VLMs.

Fig.~\ref{fig:spatial_coverage} maps the spatial distribution of GAIA samples across the globe. GAIA exhibits a wide spatial coverage, with a concentration of samples over North America, Europe, and parts of Asia. This distribution reflects the availability of high-quality satellite imagery and the focus on regions with significant human activity and environmental change. The presence of samples in diverse geographic locations ensures that models trained on GAIA can learn to generalize across various landscapes, climates, and land cover types. In addition, GAIA's stratified splits (train, test, and validation), shown in Fig.~\ref{fig:spatial_coverage}, fairly represent the global spatial coverage of the original dataset, with a balanced distribution across continents and regions. This fair representation across splits is crucial to avoid introducing biases during the training process and ensure that models can generalize well to unseen data from diverse geographical areas, while also ensures a fair, thorough, and representative assessment of model performance during testing and validation.

\begin{figure*}[h]
\centering
\includegraphics[width=\textwidth]{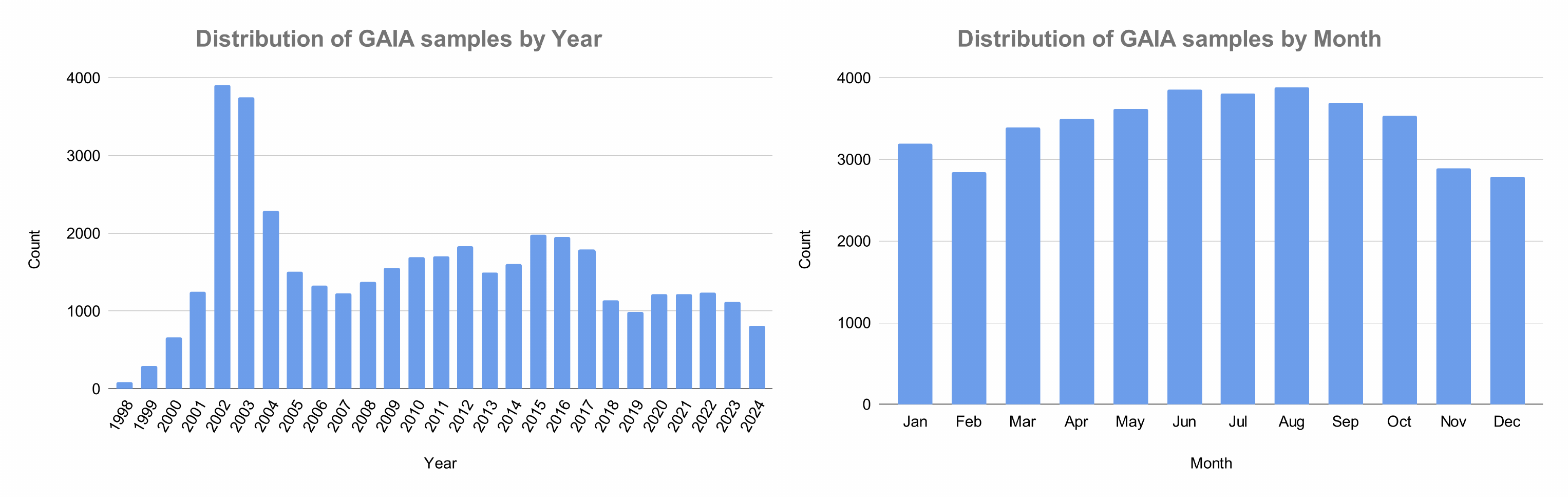}
\caption{Temporal distribution of samples within the GAIA dataset. (a) Yearly distribution of image-text pairs in GAIA, covering a period of over 25 years from 1998 to 2024. A pronounced peak in data volume is evident between 2001 and 2004. This peak coincides with the operational periods of significant satellite missions like Terra/Aqua and the increasing availability of RS data through online repositories, due to the internet uptake. A relatively uniform distribution is observed from 2008 onwards, with a slight upward trend in the last decade potentially associated with initiatives like the Copernicus Programme. (b) Monthly distribution of GAIA samples, demonstrating a relatively even representation across all months. A minor increase in sample count during the Northern Hemisphere's summer months is observed, potentially linked to reduced cloud cover and a focus on capturing seasonal phenomena such as wildfires. The consistent presence of data throughout the year ensures that GAIA captures a comprehensive range of seasonal variations.}
\label{fig:temporal_distribution}
\end{figure*}

GAIA's temporal coverage spans over 25 years, from 1998 to 2024, providing a unique perspective on Earth's changing landscapes. The distribution of samples across the years, shown in Fig.~\ref{fig:temporal_distribution} (left), reveals a peak between 2001 and 2004. This surge can be attributed to the launch and operational years of key satellite missions, such as Terra/Aqua, that are widely present in the dataset. In addition, the increasing adoption of the Internet during the early 2000s likely also contributed to this trend. The rise of online data repositories, improved data sharing capabilities, and increased awareness towards RS applications. A subsequent decrease in sample count is observed after this peak, followed by a relatively uniform distribution from 2008 onwards, with a slight increase in the last decade. This more recent trend may also be related to the establishment of large initiatives for Earth Observation data collection and distribution, such as the European Union's flagship Earth Observation Programme "Copernicus" in 2014. Fig.~\ref{fig:temporal_distribution} (right) presents the monthly distribution of GAIA samples. The data exhibits a relatively uniform distribution across the months of the year, with a slight increase observed during the summer months in the Northern Hemisphere. This may be attributed to factors such as reduced cloud cover during summer, facilitating increased image acquisition opportunities, or a potential focus on capturing seasonally prominent phenomena like wildfires. The presence of data across all months ensures representation of diverse seasonal conditions within the dataset.

\begin{figure}[h]
\centering
\includegraphics[width=\columnwidth]{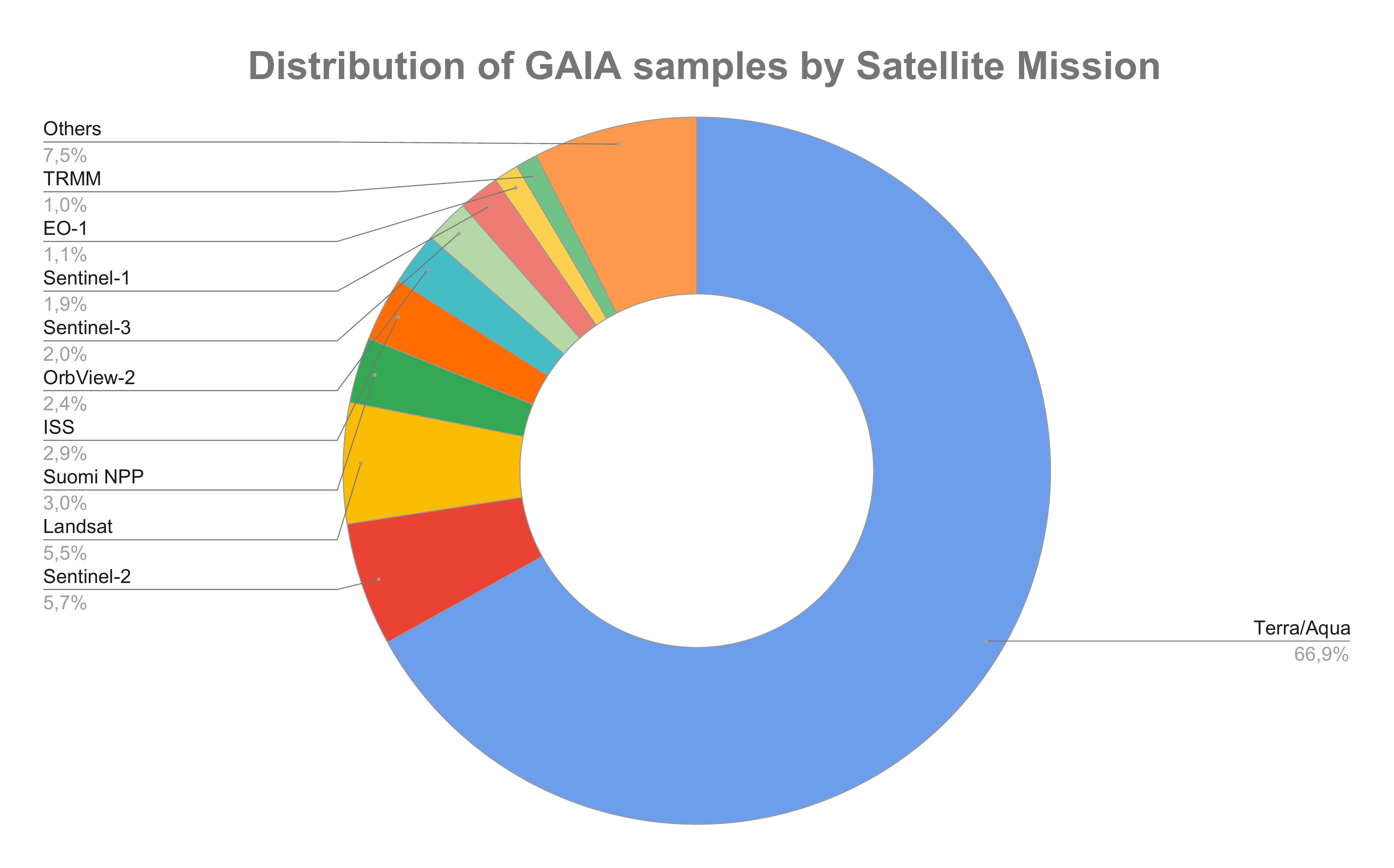}
\caption{The chart illustrates the contribution of various satellite missions to the GAIA dataset. Notably, Terra/Aqua missions are the most represented, accounting for 66.9\% of the data, attributable to their extensive operational history and broad application range. Other significant contributions come from missions such as Sentinel-2 (5.7\%), Landsat (5.5\%), and Suomi NPP (3.0\%) and the International Space Station (ISS) (2.9\%).  "Others" represents the combined contribution of the remaining satellite missions (7.5\%). All in all, the dataset includes images from over 120 satellite missions. This diversity in sensor data, including variations in spatial resolution, spectral bands, and viewing geometries, ensures a rich dataset for training and evaluating RS VLMs.}
\label{fig:satellite_distribution}
\end{figure}

Fig.~\ref{fig:satellite_distribution} shows the variety of satellite missions contributing to the GAIA dataset. In particular, GAIA includes images from more than 120 different satellite missions. The dominance of Terra/Aqua imagery (67.4\%) is evident, reflecting its long operational history and wide range of applications. However, GAIA also incorporates data from a multitude of other missions, including Sentinel-2 (5.7\%), Landsat (9.6\%) and the Suomi NPP (3.0\%), among others. Most interestingly, GAIA incorporates also images taken from the International Space Station (ISS) (3.0\%). This sensor diversity enriches the dataset with various spatial resolutions, and viewing geometries, leveraging multiple spaceborne platforms and thus enabling the development of models that are adaptable to different data sources and capable of capturing a comprehensive view of Earth's features.

\begin{figure}[h]
\centering
\includegraphics[width=\columnwidth]{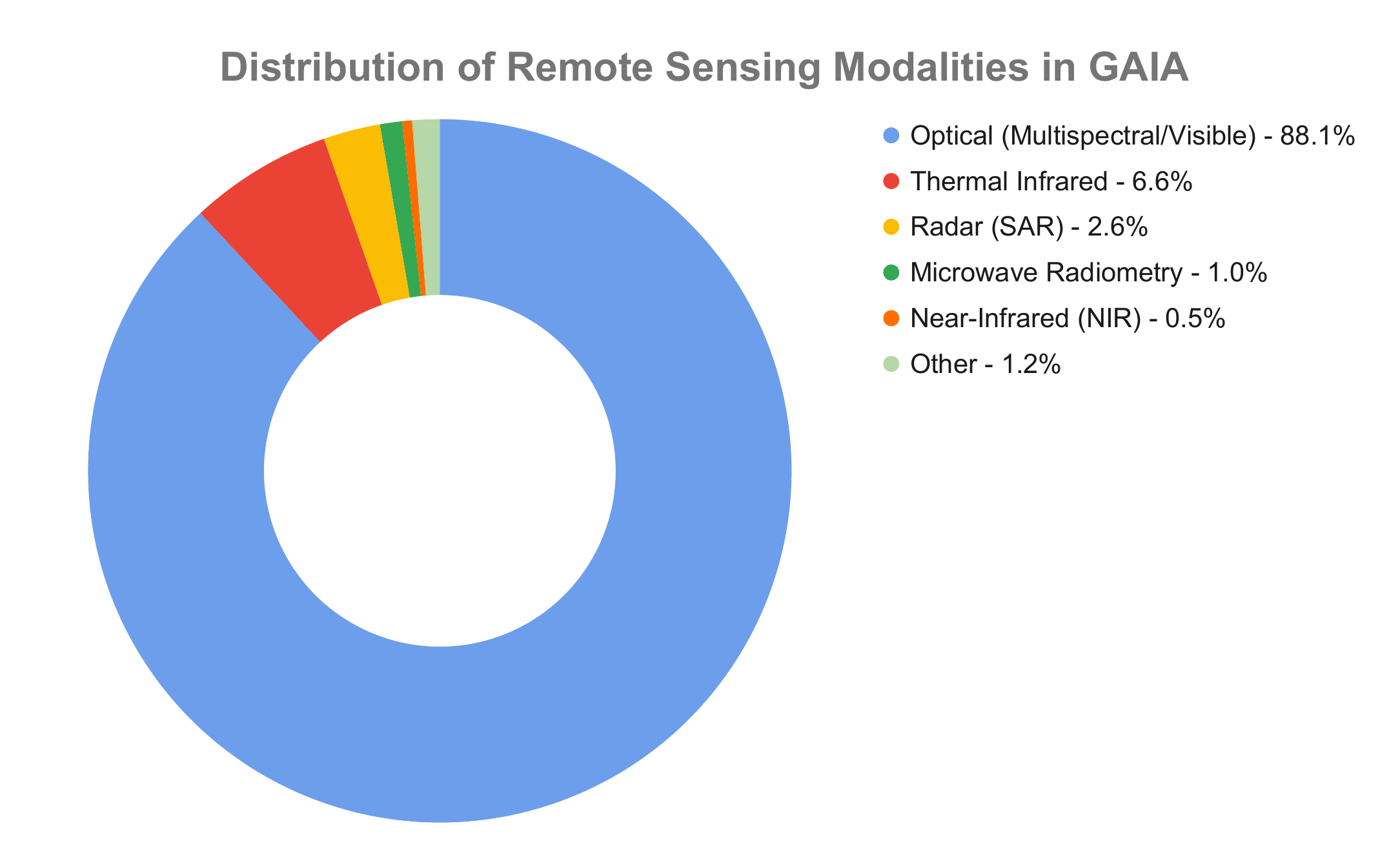}
\caption{The chart illustrates the proportion of different RS modalities represented in the GAIA dataset. Optical (multispectral/visible) imagery is the most prevalent, comprising 88.1\% of the dataset. Thermal infrared data accounts for 6.6\%, while radar (SAR) data contributes 2.6\%. Microwave radiometry and near-infrared (NIR) modalities are also present, representing 1.0\% and 0.5\% of the dataset, respectively. The "Other" category, encompassing less frequent modalities, accounts for 1.2\%. This multi-modal nature of GAIA enables the capture of a wider range of Earth phenomena beyond the visible spectrum, potentially leading to more informative and accurate image-text descriptions for various RS applications.}
\label{fig:modalities_chart}
\end{figure}

GAIA's multi-modal nature is highlighted in Fig.~\ref{fig:modalities_chart}, which displays the distribution of RS modalities. While optical/multispectral imagery constitutes the majority (88.1\%), the dataset also incorporates thermal infrared (6.6\%), radar (2.6\%), microwave radiometry (1.0\%), and near-infrared (0.5\%) data, among others. This observed distribution largely reflects the prevailing trends in RS data availability and historical usage patterns, where optical sensors have traditionally been more widely deployed and utilized, often due to factors such as the ease of acquisition and interpretability. Nevertheless, the non-optical modalities provide unique information that optical data does not capture, enabling a wider range of studies and thus enabling a more complete understanding of the Earth.

\begin{figure}[h]
\centering
\includegraphics[width=\columnwidth]{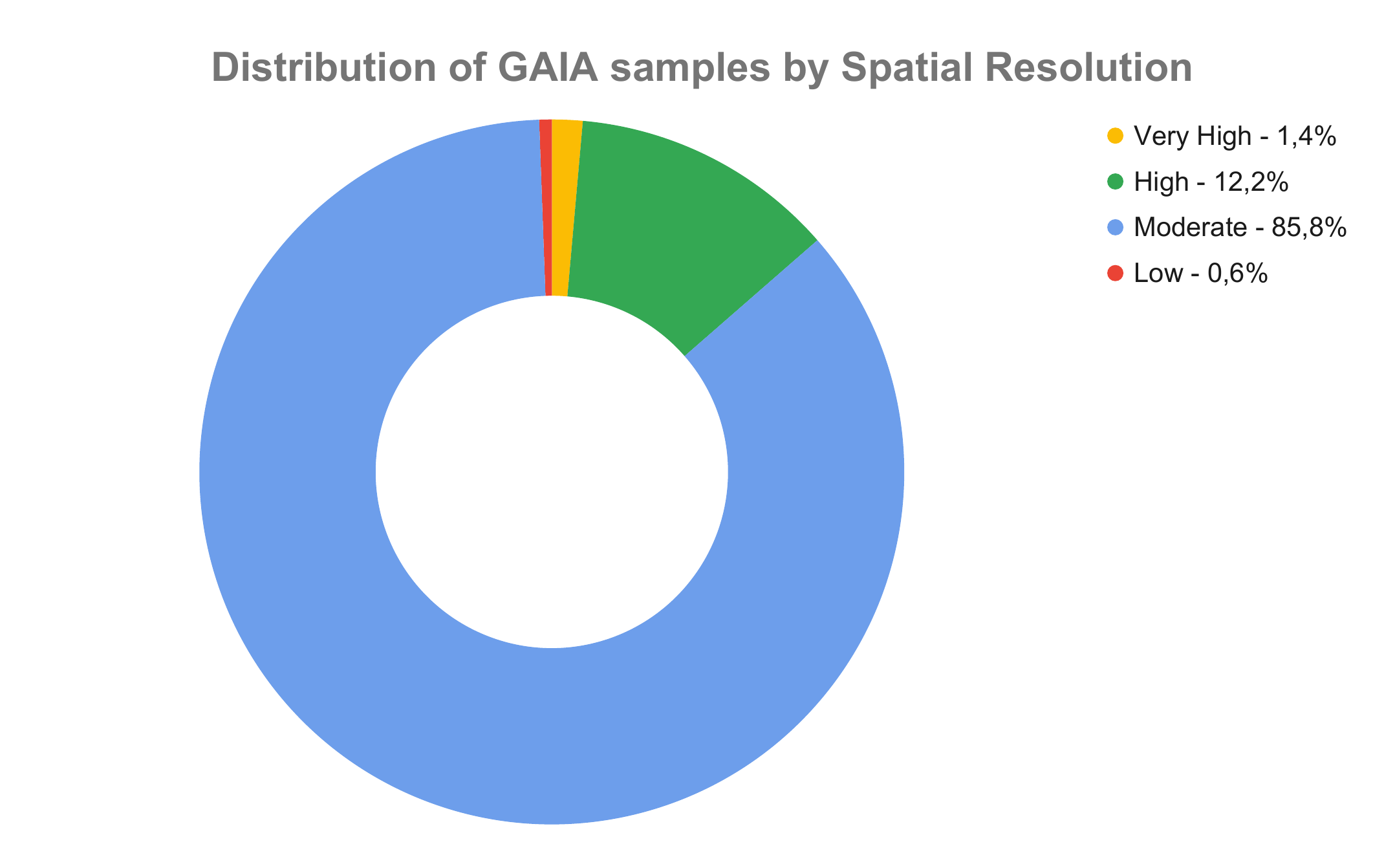}
\caption{The chart depicts the distribution of spatial resolutions within the GAIA dataset, categorized as Very High (less than 1m), High (1-10m), Moderate (10-100m), and Low (greater than 100m). Moderate-resolution imagery is the most prevalent, constituting 85.8\% of the dataset. High-resolution and Very High-resolution data account for 12.2\% and 1.4\% respectively, while Low-resolution imagery represents a smaller fraction at 0.6\%. This distribution highlights GAIA's emphasis on moderate-resolution data, which offers a balance between spatial detail and coverage area, making it suitable for a wide array of RS tasks. The inclusion of higher and lower resolution data further adds to the dataset's versatility, allowing for analyses at various scales, from broad spatial coverage to detailed feature analysis.}
\label{fig:spatial_resolution_distribution}
\end{figure}

GAIA's imagery is characterized by a diverse range of spatial resolutions, categorized as low, moderate, high, and very high. As depicted in Fig.~\ref{fig:spatial_resolution_distribution}, the majority of the samples (85.8\%) fall under the moderate resolution category. Moreover, the presence of high (12.2\%) and very high (1.4\%) resolution images allows for the exploration of fine-grained details. The negligible amount of low-resolution images (0.6\%) suggests a focus on capturing Earth's surface with sufficient detail for meaningful analysis. This plethora of diverse spatial resolution imagery makes GAIA particularly well-suited for tasks requiring either balance between broad spatial coverage or detailed feature analysis, at various scales.

\begin{figure*}[h]
\centering
\includegraphics[width=\textwidth]{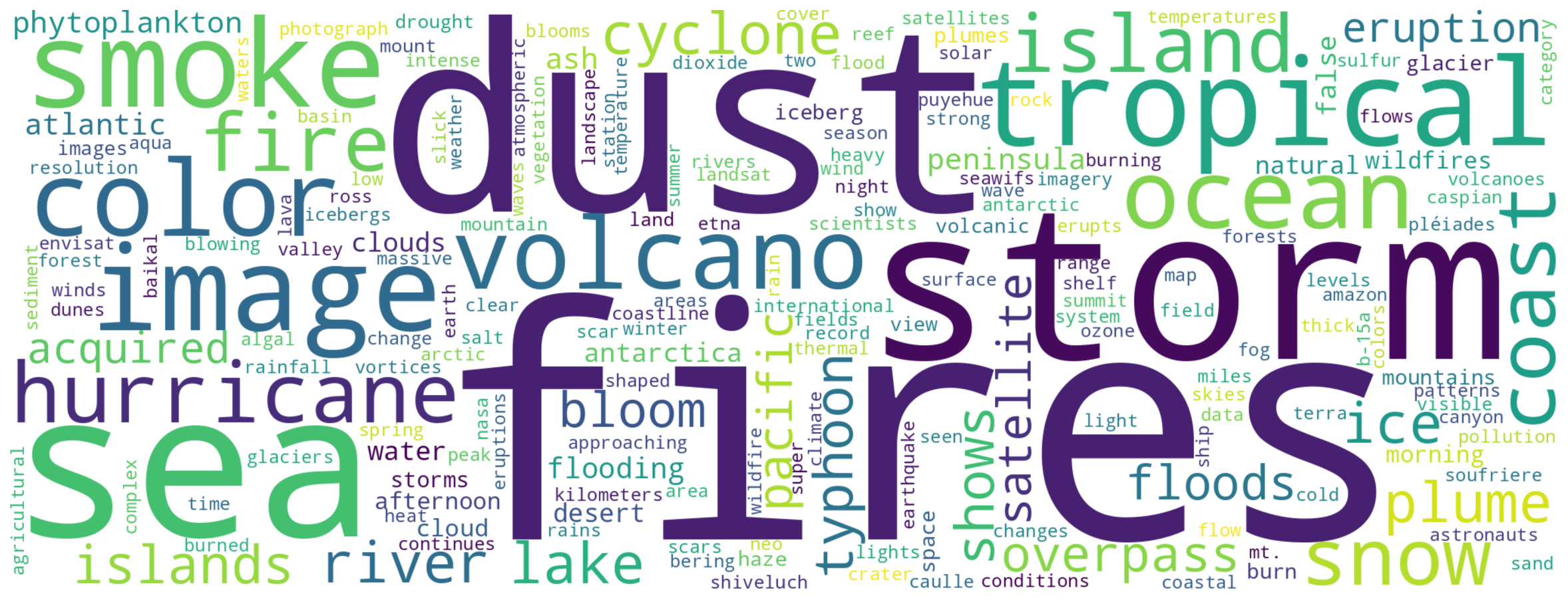}
\caption{Word cloud visualization of the most frequent terms in GAIA image descriptions. This word cloud represents the terms most commonly used to describe images in the GAIA dataset. The size of each word corresponds to its frequency of occurrence. Prominent terms such as "fires," "dust," "storm," "sea," and "volcano" highlight the dataset's focus on dynamic Earth phenomena, as well as the geographical and thematic diversity of the dataset. This visualization offers a quick overview of the key concepts and entities captured within GAIA's image descriptions.}
\label{fig:wordcloud}
\end{figure*}

Fig.~\ref{fig:wordcloud} presents a word cloud generated from the captions in the GAIA dataset, providing a glimpse into the rich semantic content captured within. Prominent terms such as "dust", "storm", "fires" and "satellite" highlight the dataset's focus on natural phenomena and RS-related concepts. The presence of geographic terms like "volcano", "island", and "ocean" underscores the diverse range of environments represented. This rich vocabulary enables the development of VLMs models that can generate detailed and contextually relevant descriptions, capturing both the physical characteristics and the underlying processes shaping our planet.

\begin{figure}[h]
\centering
\includegraphics[width=\columnwidth]{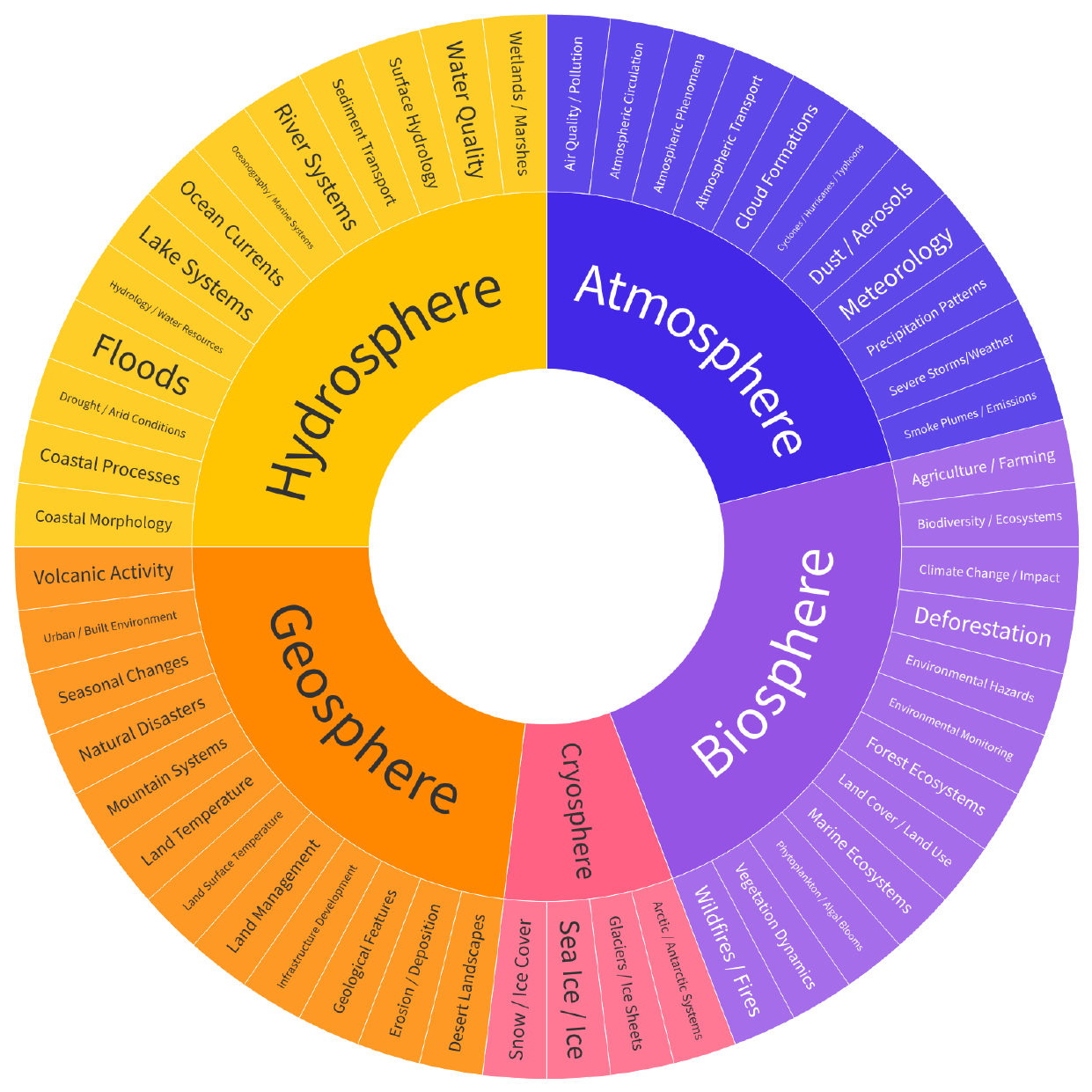}
\caption{Thematic coverage of the GAIA dataset across Earth's spheres. The GAIA dataset encompasses a broad spectrum of Earth system phenomena, categorized within the framework of the five major Earth spheres: Atmosphere, Hydrosphere, Geosphere, Biosphere, and Cryosphere. The inner ring represents these five spheres, while the outer ring segments illustrate specific subcategories of phenomena within each sphere, gathered from GAIA. This comprehensive thematic coverage demonstrates GAIA's potential to facilitate research across a multitude of Earth science disciplines, offering a holistic view of our planet's interconnected systems.}
\label{fig:categories}
\end{figure}

Fig.~\ref{fig:categories} categorizes the semantic content of GAIA into the five Earth system domains: Atmosphere, Biosphere, Cryosphere, Geosphere, and Hydrosphere. Each domain is further divided into subcategories representing specific phenomena, processes, or features based on GAIA samples thematic tags. This hierarchical organization reflects the comprehensive nature of GAIA, covering a wide spectrum of Earth science topics and providing a valuable resource for training VLMs to describe a wide array of Earth science concepts.

\begin{figure}[h]
\centering
\includegraphics[width=\columnwidth]{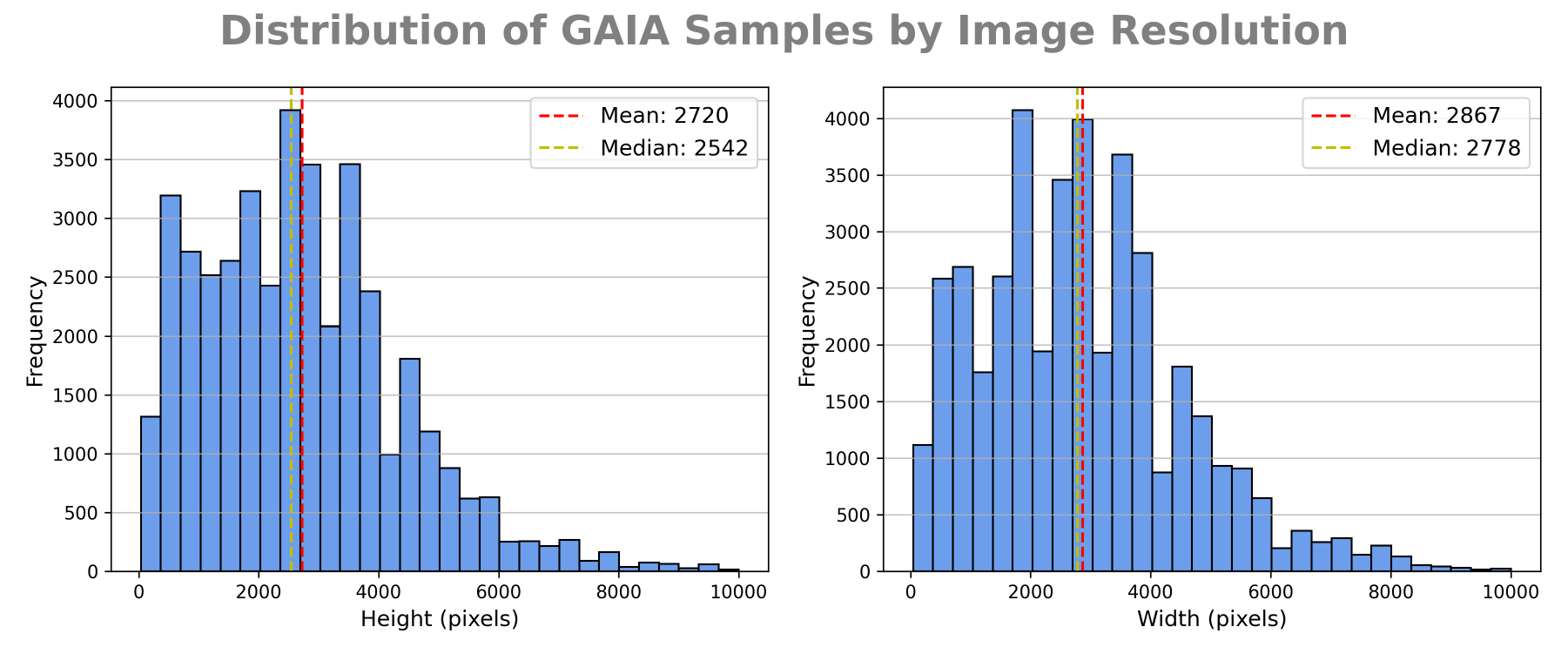}
\caption{Distribution of image dimensions (height and width) within the GAIA dataset. (a) Histogram displaying the frequency of image heights in pixels. The mean height is 2720 pixels, and the median height is 2542 pixels. (b) Histogram illustrating the frequency of image widths in pixels. The mean width is 2867 pixels, while the median width is 2778 pixels. Both distributions show a concentration of images around the 2000-4000 pixel range, indicating a prevalence of high-resolution images within the dataset, which can inform decisions regarding image processing and model development.}
\label{fig:resolutions_distribution}
\end{figure}

Finally, Fig.~\ref{fig:resolutions_distribution} presents the distribution of image heights and widths within the GAIA dataset. Both height and width distributions exhibit a bell-shaped curve, with mean values of 2720 and 2867 pixels, and median values of 2542 and 2778 pixels, respectively. These values, in conjunction with the narrow distribution around the mean indicate that GAIA images are generally large, capturing extensive spatial information. The presence of images with larger dimensions, up to 10,000 pixels, indicates variability in scene coverage and potential level of detail.

\begin{table}[h]
\centering
\caption{Comparative statistics of textual descriptions in the GAIA dataset: Alt-Text vs. Synthetic.}
\label{tab:basic_stats}
\renewcommand{\arraystretch}{1.3}
\begin{tabular}{@{}lcc@{}}
\toprule
\textbf{Attribute} & \textbf{Alt-Text} & \textbf{Synthetic} \\
\midrule
\multicolumn{3}{@{}l}{\textbf{Vocabulary}} \\
\quad Size & 15,788 & 28,396 \\
\quad Hapax Legomena & 6,403 & 7,770 \\
\quad Total Words & 379,000 & 7,346,969 \\
\midrule
\multicolumn{3}{@{}l}{\textbf{Words per Caption}} \\
\quad Minimum & 1 & 12 \\
\quad Maximum & 182 & 84 \\
\quad Average & 9.43 & 36.55 \\
\quad Median & 6.0 & 36.0 \\
\quad Standard Deviation & 12.34 & 7.20 \\
\midrule
\multicolumn{3}{@{}l}{\textbf{Sentences}} \\
\quad Avg. per Caption & 1.05 & 1.88 \\
\quad Total & 42,276 & 376,796 \\
\midrule
\multicolumn{3}{@{}l}{\textbf{Captions}} \\
\quad Total & 40,201 & 201,005 \\
\bottomrule
\end{tabular}
\end{table}

Focusing on the textual characteristics of the GAIA dataset, a comparative analysis between the original alt-text captions and our synthetic captions, shown in Table~\ref{tab:basic_stats}, reveals significant differences in their richness and complexity. The synthetic captions exhibit a considerably larger vocabulary (28,396 vs. 15,788) and a higher number of unique words (Hapax Legomena of 7,770 vs. 6,403), indicating a more diverse and nuanced use of language. This enhanced lexical diversity is further emphasized by the substantially higher total word count in the synthetic captions (7,346,969 vs. 379,000). Furthermore, the average number of words per caption is significantly greater for the synthetic (36.55) compared to the original alt-text (9.43), with a lower standard deviation (7.20 vs. 12.34), suggesting not only longer but also more consistently detailed descriptions. Each image is associated with five synthetic captions, compared to a single alt-text caption, resulting in a fivefold increase in the total number of captions and offering a diverse textual representation of each scene. 

To quantitatively assess the diversity of the five synthetic captions accompanying each image in the GAIA dataset, we conducted a comprehensive analysis. This analysis aimed to ensure that the multiple captions provide varied lexical and semantic perspectives on the visual content, thereby adding significant value over a single caption. Semantic diversity was evaluated by generating sentence embeddings using Sentence-BERT (SBERT) and calculating pairwise cosine similarities within each five-caption set. This yielded a global mean similarity of 0.7153 (median: 0.7421, std: 0.1135), corresponding to an average semantic diversity score of 0.2847 across all 40,201 sets. Concurrently, lexical diversity was evaluated using the Measure of Textual Lexical Diversity (MTLD) and Hypergeometric Distribution D (HD-D) on the combined and lemmatized text of each caption set. This achieved a mean MTLD score of 74.66 (median: 72.71, std: 16.03) and a mean HD-D score of 0.7746 (median: 0.7750, std: 0.0328). Collectively, these metrics indicate that while the five captions are semantically anchored to the same visual content, they exhibit moderate semantic diversity and notable distinctiveness in their phrasing, focus, and vocabulary richness, thereby affirming their significant value in providing multifaceted textual representations for each image.

\begin{figure}
\centering
\includegraphics[width=\columnwidth]{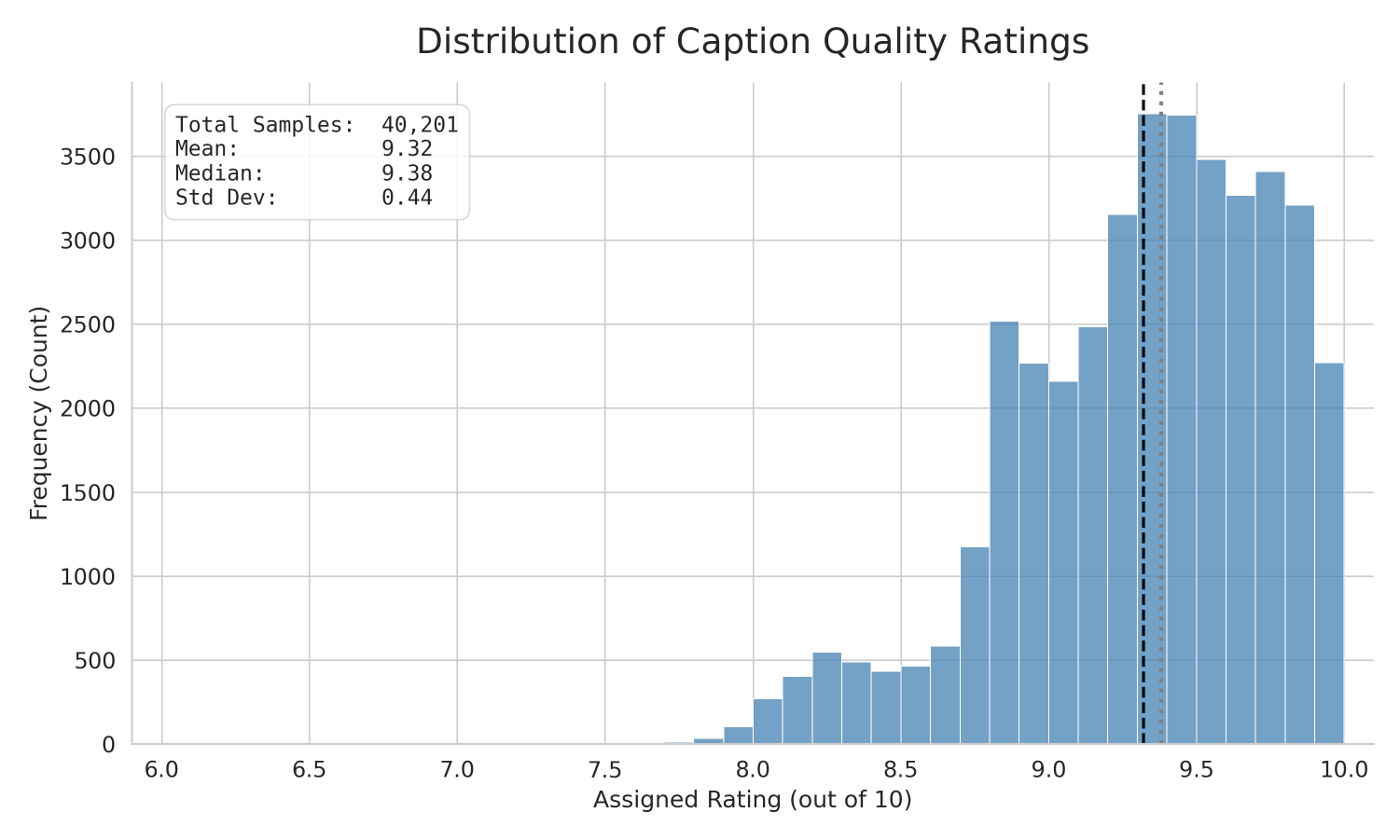}
\caption{Frequency distribution of 40,201 quality ratings assigned by gpt-4o-mini based on caption relevance and completeness. The heavily left-skewed distribution, with a mean of 9.32 (std=0.46) and a median of 9.39, provides robust quantitative evidence for our annotation pipeline's efficacy. The tight clustering of ratings at the highest end of the 10-point scale confirms our annotation pipeline's capacity to consistently generate concise, accurate, and scientifically relevant captions.}
\label{fig:caption_rating_distribution}
\end{figure}

As a final sanity check we employed gpt-4o-mini (gpt-4o-mini-2024-07-18) to verify the relevance of our generated captions with respect to the accompanied image. To this end, we prompted the model to evaluate the generated captions relevance and completeness and rate them with a score between 0 and 10 with 0 being the lowest and 10 the highest score. We end up with an overall score of 9.32 (see Fig.~\ref{fig:caption_rating_distribution}) proving the efficacy of our annotation pipeline to produce concise, accurate and scientifically relevant captions. Manual inspection of a random subset of the lower-scoring samples revealed two primary contributing factors: (a) limited context, i.e., the web-scraped text associated with the images was relatively short, and (b) constraints in the instruction-following capabilities of the model used for caption generation. While currently implemented as a retrospective validation measure, this metric possesses significant utility as an active quality-control signal for future model development. Leveraging established methodologies in robust vision-language learning, we identify three mechanisms to leverage this signal:
\begin{enumerate}
    \item \textbf{Iterative Self-Refinement:} Adopting the Self-Refine paradigm~\cite{madaan2023self}, this score functions as a reward signal within an dataset caption generation feedback loop. The model can be prompted to iteratively regenerate captions until a maximization threshold is met, ensuring outputs satisfy the strictest criteria for completeness.
    \item \textbf{Threshold-based Filtering:} Validation via CLAIR~\cite{chan2023clair} establishes LLM-based metrics as superior proxies for human judgment compared to n-gram statistics. Consequently, this score can function as a semantic gatekeeper, allowing for the pruning of relatively lower-scoring samples to curate a subset with maximized aggregate semantic fidelity.
    \item \textbf{Quality-Aware Learning:} The evaluation score can serve as a reliability index for adaptive loss weighting~\cite{kang2023noise}. By dynamically modulating the gradient contribution of each sample based on its alignment quality, the optimization process prioritizes learning from the most information-rich captions without discarding any data.
\end{enumerate}

\section{Experiments}

\subsection{Experimental Setup}
For our classification and cross-modal retrieval experiments, we utilized CLIP~\cite{radford2021learning} pre-trained vision transformer (ViT) models~\cite{dosovitskiy2020image}. The models were loaded using the OpenCLIP library~\cite{ilharco_gabriel_2021_5143773}. Our evaluation methodology followed the setup described in~\cite{radford2021learning} and utilized the CLIP Benchmark~\footnote{\url{https://github.com/LAION-AI/CLIP_benchmark}} evaluation suite to ensure a consistent and fair comparison with existing zero-shot CLIP benchmarks. Classification performance was assessed using Accuracy (Acc1 and Acc5) and Mean Recall, while the retrieval task performance was evaluated using Recall@K (R@K).

For our image captioning experiments, we employed the BLIP2~\cite{li2023blip} pre-trained model, specifically opting for its smallest and most resource-efficient variant, the blip2-opt-2.7b model. This choice allowed us to focus on fine-tuning the model exclusively using our image-caption paired dataset, eliminating the need for additional instruction tuning datasets, as required by models like LLaVA~\cite{li2023stablellava,li2024llava}. To evaluate image captioning performance, we employed commonly used metrics, including BLEU~\cite{papineni2002bleu}, ROUGE-1, ROUGE-2, ROUGE-L~\cite{lin2004rouge}, and METEOR~\cite{banerjee2005meteor}, ensuring a comprehensive assessment of the model's capabilities.

We created three dataset splits (train, test, and validation - see Fig.~\ref{fig:spatial_coverage}) with a 70/20/10 ratio, spatio-temporally stratified. Temporal stratification was based on grouping GAIA samples into seasons according to their acquisition date. Spatial stratification was achieved using Geohash~\cite{niemeyer2008geohash}, a hierarchical geocoding system that partitions the Earth's surface into a grid of cells. A Geohash precision of 3 was used, corresponding to a grid cell size of approximately $156km\times156km$. The stratified global spatial and temporal coverage of these splits is illustrated in Fig.~\ref{fig:spatial_coverage}.

Fine-tuning of both the CLIP models and the BLIP2 was performed on a single Nvidia GeForce RTX 4090 GPU. For the CLIP models fine-tuning, the predefined train split of our dataset was used with a batch size of 128 for a single epoch. For the single alt-text captions we perform fine-tuning as usual, while for the five synthetic captions, we incorporate an intermediate step where we average the embeddings of all five captions to produce a unified representation. We employed a learning rate of 5e-6 with a 10\% warmup ratio, a cosine annealing learning rate schedule, and the AdamW optimizer~\cite{kingma2014adam, loshchilov2017decoupled} with a weight decay of 0.1~\cite{andriushchenko2023need}. Gradient checkpointing~\cite{chen2016training} was enabled for the largest models to accommodate memory constraints without reducing the batch size. No data augmentation techniques were employed. During fine-tuning, both the vision and text encoders remain trainable, and no additional trainable parameters were introduced to the models. For BLIP-2 model fine-tuning, we adopted a parameter-efficient approach using Low-Rank Adaptation (LoRA)~\cite{hu2021lora}. The model was fine-tuned on the predefined training split of our dataset using a batch size of 16 over five epochs. LoRA was applied to the attention layers with a rank of 16, a scaling factor $\alpha$ of 32, and a dropout rate of 0.1. Unlike our CLIP fine-tuning methodology, which averages the text embeddings from the five synthetic captions to generate a unified text representation, we adopted a sampling approach for BLIP-2. During fine-tuning, a single caption was randomly selected per image at each epoch. This strategy preserves caption diversity while mitigating overfitting risks. Similarly to the CLIP setup, no data augmentation was applied.

Ultimately, we assessed classification performance on three benchmark datasets featured in the CLIP~\cite{radford2021learning} benchmark suite, enabling direct comparisons across datasets: (1) two RS scene classification datasets, EuroSAT and RESISC45, and (2) ImageNet1K, a well-established general-domain benchmark where CLIP models typically achieve strong performance. RS scene classification, is one of the few tasks presented by the CLIP authors, where zero-shot CLIP significantly under-performs on the EuroSAT and RESISC45 datasets, resulting in some of the biggest deltas (-37,1\% and -11.9\% respectively) among all tasks assessed, when compared to a fully supervised ResNet50 baseline model. By including ImageNet-1K, we aimed to investigate the impact of GAIA fine-tuning on CLIP's retained capabilities in tasks where its baseline performance is already robust. For the evaluation of cross-modal retrieval and image captioning, we employed the GAIA test split. We performed a comparative analysis of cross-modal alignment performance, assessing the performance both before and after fine-tuning, and contrasting the outcomes for alt-text captions against those obtained using synthetic captions. This allowed us to quantify the performance gains associated with each captioning approach. Furthermore, for image captioning assessment, we evaluated two key aspects. We first quantified the performance improvement resulting from fine-tuning compared to zero-shot performance. Secondly, we analyzed the performance enhancements associated with each individual captioning scheme.

\subsection{Experimental Results}
We present the results of our CLIP experiments in Table~\ref{tab:classification} and Table~\ref{tab:retrieval}, for classification and retrieval, respectively. Both tables highlight the benefits of fine-tuning CLIP models with GAIA, i.e. using semantically rich image-text paired datasets, for both classification and cross-modal retrieval. More specifically, Table~\ref{tab:classification} reveals that fine-tuning, caption quality, and model scale emerge as key factors significantly influencing CLIP classification performance. ImageNet1k demonstrates the highest performance, while RS datasets benefit the most from fine-tuning, indicating the need for domain adaptation. Synthetic captions have a positive impact on classification, providing substantial improvements compared to alt-text captions. Significant Acc1, Acc5, and Mean Recall gains are observed post fine-tuning, especially for EuroSAT, addressing the known zero-shot CLIP under-performance in RS. For RESISC45 on the other hand, while fine-tuning influences the vast majority of the 45 dataset classes positively, in terms of evaluation metrics, it affects negatively classes such as "Airplane", "Airport", "Church", leading to marginal performance gains overall. These modest performance gains could be attributed to the GAIA dataset thematics as well as the limited representation of "very high" and "high" spatial resolution imagery in the dataset, as shown in Fig.~\ref{fig:spatial_resolution_distribution}. Importantly, fine-tuning on GAIA does not degrade ImageNet1k performance and may even slightly improve it, suggesting positive transfer learning. Lastly, larger models generally perform better in classification, which is consistent with scaling trends. 

Table~\ref{tab:retrieval} reveals significant enhancements in cross-modal retrieval after fine-tuning and strongly underscores the necessity of fine-tuning and the value of high-quality captions for effective cross-modal retrieval in specialized domains. Across all model sizes and caption types, fine-tuned models consistently demonstrate substantially improved Recall@K metrics in both text-to-image and image-to-text retrieval. Synthetic captions consistently outperform Alt-text captions, indicating that richer, semantically nuanced text provides better cross-modal alignment. This advantage is evident across all R@K levels and model sizes.  As expected, larger CLIP models (ViT-L-14, ViT-L-14-336) consistently outperform the smaller ViT-B-32, demonstrating the benefit of increased model capacity for these tasks.  The performance gap between fine-tuned and zero-shot models, and between caption types, widens at higher K values, suggesting improved overall ranking quality.

For completeness, we compare both our alt-text and synthetic captions fine-tuned CLIP-ViT-L-14 models with the zero-shot CLIP model and other RS-domain fine-tuned CLIP models, i.e. SkyCLIP~\cite{wang2024skyscript}, RemoteCLIP~\cite{liu2024remoteclip}, GeoRSCLIP~\cite{zhang2024rs5m}, LAION\_RS~\cite{wang2024skyscript}. Despite using a dataset several times smaller than those employed to fine-tune the rest of the models considered and training for only a single epoch, GAIA delivers strong classification and retrieval performance.

Table~\ref{tab:rs_clip_comparison_cls} reveals that for the classification task, GAIA (Synthetic) ranks among the top-performing models. On EuroSAT and ImageNet1K, our GAIA (Synthetic) model positions itself among the top-performing methods. For EuroSAT, it achieves an Acc1 of 72.46\%, outperforming SkyCLIP (70.54\%) and LAION\_RS (71.33\%), and is competitive with RS5M (74.52\%). On ImageNet1K, its Acc1 of 75.49\% is on par with SkyCLIP and the baseline CLIP, and approaches LAION\_RS (75.61\%), demonstrating strong generalization to common vision tasks. Conversely, on RESISC45, the performance of our GAIA-tuned models is closer to the zero-shot CLIP baseline. This discrepancy may be attributed to the GAIA's comparatively limited inclusion of high-resolution aerial imagery, a characteristic of RESISC45, as discussed in our exploratory data analysis (see Section~\ref{sec:eda}). Nevertheless, the overall results underscore GAIA's potential for efficient, effective and semantically rich fine-tuning, yielding competitive performance with significantly reduced data and computational requirements.

On the other hand, for cross-modal retrieval, in order to better demonstrate GAIA’s unique features through RS tailored benchmarks, we employed the thematic tags information associated with each GAIA sample. More specifically, each GAIA sample is complemented by three generic tags, which encompass thematic aspects such as earth sphere, topic, and location (e.g., ["Cryosphere", "Climate Change", "Arctic"] or ["Floods", "Sediment Transport", "Adriatic Sea"]). We deliberately chose to base our benchmarks on these tags rather than on the aggregated five major earth spheres and their subcategories, shown in Fig.~\ref{fig:categories}. This choice was motivated by two key considerations:
\begin{enumerate}
    \item \textbf{Higher Semantic Variance:} The thematic tags provide richer semantic information compared to the earth sphere aggregates (5) and its subcategories which are relatively coarse and limited in variability.
    \item \textbf{Relationship Structure:} Due to the fact that benchmarks and metrics commonly used in the context of CLIP, rely on one-to-one relationships for evaluation, GAIA's thematic tags present a more challenging and realistic setting to validate models capacity for semantic understanding and cross-modal alignment. More specifically, the tags form complex one-to-one matches with images (>6600 different thematic tag combinations within our GAIA test split), whereas the earth spheres (5) and subcategories typically yield simpler one-to-many mappings.
\end{enumerate}
\noindent Table~\ref{tab:rs_clip_comparison_retrieval} demonstrates that our GAIA-fine-tuned models, particularly GAIA (Synthetic), establish a new state-of-the-art across all recall metrics when evaluated using GAIA's thematic tags. For text-to-image retrieval, the GAIA (Synthetic) model achieves a Recall@100 of 47.32\%, outperforming the zero-shot CLIP baseline (38.07\%) and the second-best model, LAION\_RS (45.03\%). The advantage is even more pronounced in image-to-text retrieval, where GAIA (Synthetic) reaches a Recall@100 of 51.79\%, creating a large margin over both the baseline CLIP (36.97\%) and the rest of RS-domain fine-tuned models. Finally, our GAIA (Alt-text) model also delivers very strong performance, exhibiting the second-best results in image-to-text retrieval and third-best results in text-to-image retrieval.

We evaluate the performance of the BLIP2 captioning model when fine-tuned on the GAIA dataset, using two different captioning schemes: original alt-text captions and synthetic captions. We analyze both quantitative metrics and qualitative examples to understand the capabilities and limitations of the fine-tuned models. The evaluation metrics, summarized in Table~\ref{tab:captioning}, reveal a clear improvement in captioning performance after fine-tuning BLIP2 on the GAIA dataset. Both fine-tuning approaches, using either alt-text or synthetic captions, significantly outperform the zero-shot BLIP2 model across all metrics (BLEU, ROUGE-1, ROUGE-2, ROUGE-L, METEOR). Fine-tuning substantially improves performance. Models fine-tuned on GAIA's synthetic captions consistently outperform those fine-tuned on alt-text captions. This suggests that the richer and more semantically meaningful synthetic captions provide a better training signal in order for BLIP2 to learn to produce more nuanced in-domain descriptions. The most significant improvements are observed in ROUGE and METEOR scores, which are more sensitive to semantic similarity and recall, while BLEU, focused on precision and n-gram overlap, shows a lower absolute improvement but still a notable relative gain. This indicates that fine-tuning helps BLIP2 generate captions that are more semantically similar to the reference captions, although n-gram overlap (BLEU) might still be relatively limited. Overall, the quantitative metrics strongly indicate that fine-tuning BLIP2 on GAIA dataset, especially using synthetic captions, significantly enhances its RS image captioning capabilities. 

\begin{figure*}
\centering
\includegraphics[width=0.8\textwidth]{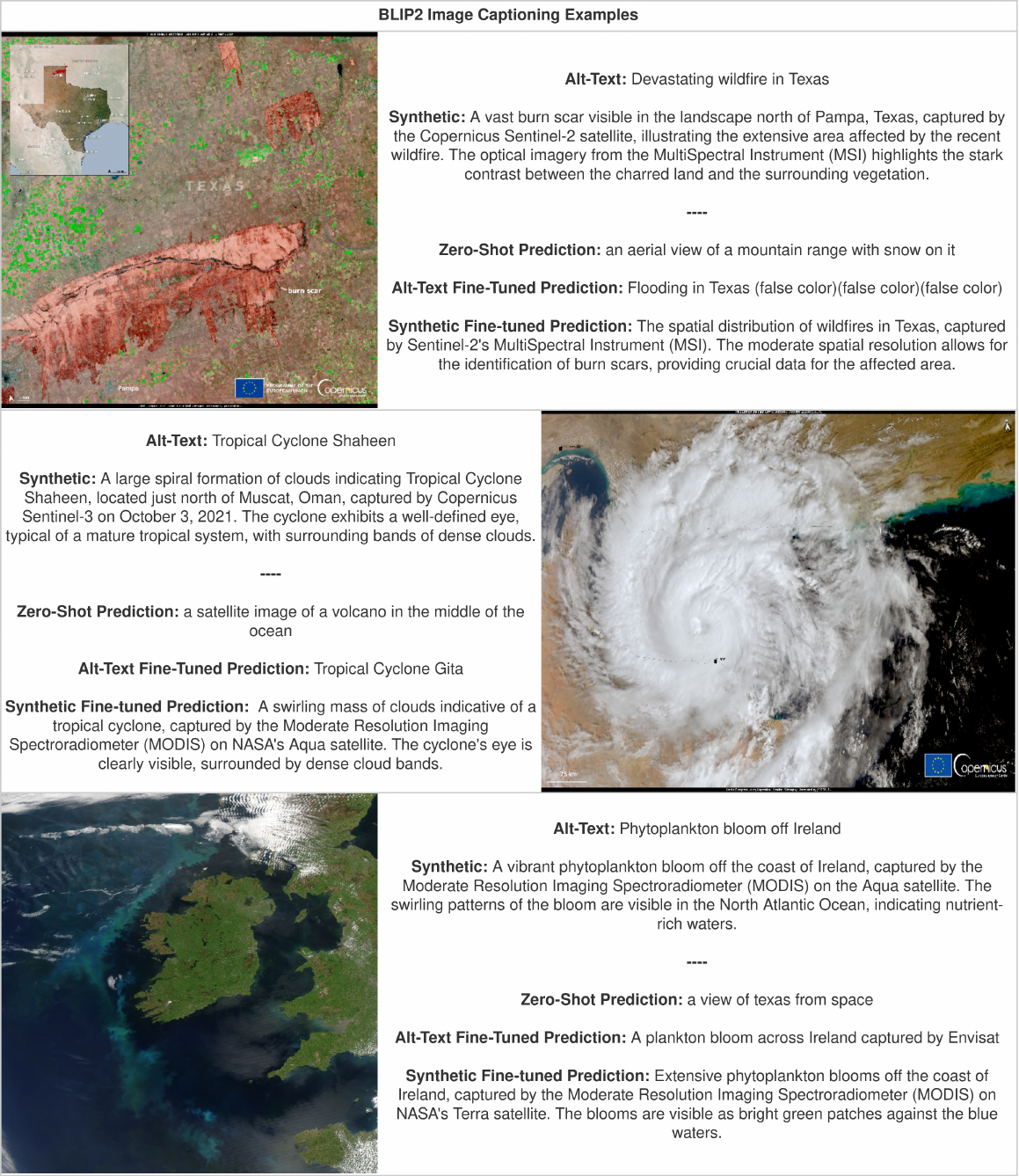}
\caption{GAIA image captioning examples using BLIP2. This figure presents three diverse satellite images featuring diverse Earth Observation scenarios. For each image, we illustrate the corresponding Alt-Text and Synthetic captions along with the zero-shot, alt-text fine-tuned and synthetic fine-tuned predictions, demonstrating the semantic difference between the alt-text and synthetic captions as well as the evolution of captioning capabilities.}
\label{fig:blip_samples}
\end{figure*}

A qualitative analysis of the generated captions (see Fig.~\ref{fig:blip_samples}) provides deeper insights into the fine-tuned model's strengths and weaknesses, which might be traced back to GAIA and reveal potential limitations or potent areas for improvement. Fine-tuned models, particularly those trained on synthetic captions, generate predictions that are more relevant to the image content and reference captions compared to zero-shot predictions. They are better at identifying key elements, locations, and events described in the references captions. Moreover, synthetic caption fine-tuned models show improved ability to handle complex RS scenes, such as those involving natural disasters (wildfires, floods, cyclones) and environmental phenomena (phytoplankton blooms, dust storms). They more accurately identify these events compared to their other two counterparts. Last but not least, models fine-tuned using our synthetic captions generate prominently more extensive and informative descriptions.

Despite those improvements, fine-tuned models occasionally hallucinate or generate inaccurate details. The models sometimes misidentify locations, confusing geographically proximate but distinct places. This suggests a challenge in precisely grounding the captions geographically, even after fine-tuning. Moreover, while the models generally improve in correctly identifying the broad modality of RS imagery (e.g., optical, radar), they still confuse specific sensors and missions. For instance, a caption might misidentify "Sentinel-2" for "MODIS" or most trivially correctly identify "MODIS" but incorrectly attribute it to "Terra" instead of "Aqua," or vice versa. Lastly, models fine-tuned on synthetic captions, while generating more informative descriptions, sometimes produce captions that are excessively verbose or redundant, indicating a need for better control over caption length and conciseness.

Lastly, we further evaluate our best-performing models - the BLIP2 and CLIP-ViT-L-14 variants fine-tuned on GAIA's synthetic captions - on two additional RS image-text paired datasets: RSICap and GIT-10M. In the absence of standard evaluation splits for GIT-10M, we curated a stratified test set of 100k samples from its georeferenced partition. Adopting the GAIA spatial stratification protocol, we maximized global geographic diversity while simultaneously balancing the dataset across a Ground Sample Distance (GSD) spectrum of 0.5m to 128m to ensure scale invariance. The resulting subset exhibits significant linguistic variance, characterized by a mean caption length of 52 words (min=7, max=127). The results, presented in Table~\ref{tab:captioning_retrieval_combined}, summarize the performance of our GAIA fine-tuned models against their zero-shot baselines. For the cross-modal retrieval task, our CLIP-ViT-L-14 model exhibits significant improvements. For both text-to-image and image-to-text recall, our fine-tuned model consistently outperforms the zero-shot baseline across all Recall@k metrics. As for image captioning the results demonstrate that the BLIP2 model fine-tuned on GAIA achieves substantial performance gains on both external datasets. On RSICap, the METEOR score doubles from 7.35 to 15.09, and the ROUGE-L score increases from 15.21 to 18.76. The improvement is even more pronounced on GIT-10M, where the METEOR score jumps from 6.83 to 16.62. Collectively, these results demonstrate that the knowledge encoded in the GAIA dataset equips models with robust, generalizable knowledge that significantly enhances performance on a diverse range of RS downstream tasks. However, it is important to contextualize these improvements by highlighting the significant annotation differences between GAIA's captions and those of other RS image-text paired datasets (see Fig.~\ref{fig:vlm_datasets}). While the performance gains are clear, reference-based metrics may not fully capture the semantic quality of the generated captions, as they are being compared solely against reference captions from datasets exhibiting different annotation paradigms. Typically, fine-tuning induces a linguistic bias that degrades performance on datasets with different annotation style. However, our analysis attributes the observed gains to two countervailing mechanisms:
\begin{enumerate}
    \item \textbf{Visual Feature Adaptation:} The zero-shot BLIP-2 baseline has been pre-trained primarily on natural, web-crawled imagery, resulting in a significant domain gap when processing satellite/aerial perspectives. Fine-tuning on GAIA allows the model to learn these overhead visual features and subsequently relate their semantic content to domain-specific vocabulary. This ability to correctly identify objects is a fundamental improvement that applies to any dataset, regardless of the caption style.
    \item \textbf{Compensating for Metric Penalty:} Conventional evaluation metrics rely heavily on n-gram overlap and inherently penalize stylistic deviations from the ground truth. Since GAIA induces a linguistic shift distinct from the styles of RSICap or GIT-10M, a degradation in these metrics would be the theoretical expectation. The fact that quantitative scores increased suggests that the gain in semantic fidelity, i.e. the ability to correctly infer content and utilize precise terminology, is substantial enough to outweigh the penalty imposed by the stylistic mismatch.
\end{enumerate}
\noindent Last but not least, we must highlight the stark differences in terms of both visual feature recognition competence and the level of description detail and domain-specific vocabulary use of the zero-shot BLIP2 model prediction, observed in Fig.~\ref{fig:blip_samples}.

\begin{table*}[htbp]
\caption{Classification Results Grouped by Model and Caption Scheme pre and post fine-tuning with either of the GAIA caption schemes}
\label{tab:classification}
\centering
\scriptsize
\sisetup{table-format=2.2, table-number-alignment=center}
\begin{tabular}{@{}lllccSSS@{}}
\toprule
\multicolumn{1}{c}{\multirow{2}{*}{\centering Dataset}} & 
\multicolumn{1}{c}{\multirow{2}{*}{\centering Model}} & 
\multicolumn{1}{c}{\multirow{2}{*}{\centering Captions}} & 
\multicolumn{1}{c}{\multirow{2}{*}{\centering Mode}} & 
\multicolumn{3}{c}{Metrics (\%)} \\
\cmidrule(lr){5-7}
 & & & & {Acc1} & {Acc5} & {Mean Recall} \\
\midrule

% ================ eurosat ================
\multirow{12}{*}{\centering\rotatebox[origin=c]{90}{eurosat}} 
& \multirow{4}{*}{\centering ViT-B-32} 
& \multirow{2}{*}{\centering Alt-text} & Zero-shot & 50.46 & 92.41 & 49.06 \\
& & & Fine-tuned & 61.48 & 96.68 & 60.41 \\
\cmidrule(lr){3-7}
& & \multirow{2}{*}{\centering Synthetic} & Zero-shot & 50.46 & 92.41 & 49.06 \\
& & & Fine-tuned & 62.00 & 95.50 & 61.12 \\
\cmidrule(lr){2-7}

& \multirow{4}{*}{\centering ViT-L-14} 
& \multirow{2}{*}{\centering Alt-text} & Zero-shot & 62.57 & 95.93 & 63.95 \\
& & & Fine-tuned & 68.52 & 98.87 & 69.75 \\
\cmidrule(lr){3-7}
& & \multirow{2}{*}{\centering Synthetic} & Zero-shot & 62.57 & 95.93 & 63.95 \\
& & & Fine-tuned & 72.46 & 98.91 & 72.68 \\
\cmidrule(lr){2-7}

& \multirow{4}{*}{\centering ViT-L-14-336} 
& \multirow{2}{*}{\centering Alt-text} & Zero-shot & 61.52 & 96.06 & 62.97 \\
& & & Fine-tuned & 69.02 & 98.89 & 70.62 \\
\cmidrule(lr){3-7}
& & \multirow{2}{*}{\centering Synthetic} & Zero-shot & 61.52 & 96.06 & 62.97 \\
& & & Fine-tuned & 74.26 & 99.18 & 74.11 \\
\midrule

% ================ resisc45 ================
\multirow{12}{*}{\centering\rotatebox[origin=c]{90}{resisc45}} 
& \multirow{4}{*}{\centering ViT-B-32} 
& \multirow{2}{*}{\centering Alt-text} & Zero-shot & 53.65 & 86.71 & 54.08 \\
& & & Fine-tuned & 52.11 & 84.84 & 52.46 \\
\cmidrule(lr){3-7}
& & \multirow{2}{*}{\centering Synthetic} & Zero-shot & 53.65 & 86.71 & 54.08 \\
& & & Fine-tuned & 56.54 & 86.81 & 56.27 \\
\cmidrule(lr){2-7}

& \multirow{4}{*}{\centering ViT-L-14} 
& \multirow{2}{*}{\centering Alt-text} & Zero-shot & 63.35 & 92.40 & 63.82 \\
& & & Fine-tuned & 65.49 & 91.95 & 66.01 \\
\cmidrule(lr){3-7}
& & \multirow{2}{*}{\centering Synthetic} & Zero-shot & 63.35 & 92.40 & 63.82 \\
& & & Fine-tuned & 66.11 & 92.65 & 66.62 \\
\cmidrule(lr){2-7}

& \multirow{4}{*}{\centering ViT-L-14-336} 
& \multirow{2}{*}{\centering Alt-text} & Zero-shot & 63.73 & 92.73 & 64.31 \\
& & & Fine-tuned & 65.83 & 92.48 & 66.28 \\
\cmidrule(lr){3-7}
& & \multirow{2}{*}{\centering Synthetic} & Zero-shot & 63.73 & 92.73 & 64.31 \\
& & & Fine-tuned & 65.98 & 92.79 & 66.45 \\
\midrule

% ================ imagenet1k ================
\multirow{12}{*}{\centering\rotatebox[origin=c]{90}{imagenet1k}} 
& \multirow{4}{*}{\centering ViT-B-32} 
& \multirow{2}{*}{\centering Alt-text} & Zero-shot & 63.33 & 88.81 & 63.35 \\
& & & Fine-tuned & 59.76 & 86.16 & 59.74 \\
\cmidrule(lr){3-7}
& & \multirow{2}{*}{\centering Synthetic} & Zero-shot & 63.33 & 88.81 & 63.35 \\
& & & Fine-tuned & 61.61 & 87.52 & 61.62 \\
\cmidrule(lr){2-7}

& \multirow{4}{*}{\centering ViT-L-14} 
& \multirow{2}{*}{\centering Alt-text} & Zero-shot & 75.54 & 94.58 & 75.54 \\
& & & Fine-tuned & 75.32 & 94.66 & 75.33 \\
\cmidrule(lr){3-7}
& & \multirow{2}{*}{\centering Synthetic} & Zero-shot & 75.54 & 94.58 & 75.54 \\
& & & Fine-tuned & 75.49 & 94.57 & 75.48 \\
\cmidrule(lr){2-7}

& \multirow{4}{*}{\centering ViT-L-14-336} 
& \multirow{2}{*}{\centering Alt-text} & Zero-shot & 76.57 & 95.12 & 76.55 \\
& & & Fine-tuned & 76.34 & 95.11 & 76.35 \\
\cmidrule(lr){3-7}
& & \multirow{2}{*}{\centering Synthetic} & Zero-shot & 76.57 & 95.12 & 76.55 \\
& & & Fine-tuned & 76.48 & 95.12 & 76.48 \\
\bottomrule
\end{tabular}
\end{table*}

\begin{table*}[htbp]
\caption{Cross-Modal Retrieval Results Grouped by Model and GAIA Caption Scheme}
\label{tab:retrieval}
\centering
\sisetup{table-format=2.2, table-number-alignment=center}
\scriptsize
\begin{tabular}{@{}lllcccccccc@{}}
\toprule
\multicolumn{1}{c}{\multirow{2}{*}{Model}} & 
\multicolumn{1}{c}{\multirow{2}{*}{Captions}} & 
\multicolumn{1}{c}{\multirow{2}{*}{Mode}} & 
\multicolumn{4}{c}{Text-to-Image Recall (\%)} & 
\multicolumn{4}{c}{Image-to-Text Recall (\%)} \\
\cmidrule(lr){4-7} \cmidrule(lr){8-11}
& & & @1 & @5 & @10 & @20 & @1 & @5 & @10 & @20 \\
\midrule

% =============== ViT-B-32 ===============
\multirow{4}{*}{\centering ViT-B-32} 
& \multirow{2}{*}{Alt-Text} 
& Zero-shot & 3.13 & 9.66 & 15.08 & 22.44 & 2.39 & 6.77 & 10.53 & 15.96 \\
& & Fine-tuned & 6.57 & 19.60 & 29.33 & 41.18 & 5.99 & 19.01 & 28.77 & 40.64 \\
\cmidrule(lr){2-11}
& \multirow{2}{*}{Synthetic} 
& Zero-shot & 4.89 & 13.49 & 20.03 & 28.63 & 4.89 & 13.96 & 20.59 & 29.35 \\
& & Fine-tuned & 10.37 & 29.38 & 40.47 & 55.22 & 10.72 & 29.72 & 41.46 & 54.89 \\
\midrule

% =============== ViT-L-14 ===============
\multirow{4}{*}{\centering ViT-L-14} 
& \multirow{2}{*}{Alt-Text} 
& Zero-shot & 7.92 & 20.63 & 30.04 & 41.34 & 5.49 & 13.94 & 20.64 & 28.78 \\
& & Fine-tuned & 11.16 & 29.78 & 41.47 & 55.34 & 09.72 & 28.05 & 39.65 & 53.29 \\
\cmidrule(lr){2-11}
& \multirow{2}{*}{Synthetic} 
& Zero-shot & 12.69 & 29.46 & 39.71 & 51.38 & 12.24 & 29.10 & 39.20 & 51.18 \\
& & Fine-tuned & 20.56 & 45.81 & 59.09 & 72.72 & 19.90 & 44.69 & 58.49 & 71.87 \\
\midrule

% ============= ViT-L-14-336 =============
\multirow{4}{*}{\centering ViT-L-14-336} 
& \multirow{2}{*}{Alt-Text} 
& Zero-shot & 8.87 & 23.59 & 33.57 & 45.24 & 6.34 & 16.34 & 23.27 & 32.75 \\
& & Fine-tuned & 12.66 & 33.22 & 44.92 & 58.99 & 11.35 & 31.75 & 43.58 & 57.94 \\
\cmidrule(lr){2-11}
& \multirow{2}{*}{Synthetic} 
& Zero-shot & 14.59 & 33.62 & 44.66 & 56.89 & 15.22 & 33.68 & 44.15 & 55.74 \\
& & Fine-tuned & 22.92 & 48.91 & 62.39 & 75.83 & 22.66 & 48.46 & 61.53 & 75.09 \\
\bottomrule
\end{tabular}
\end{table*}

\begin{table*}[ht]
\caption{Remote Sensing Image Captioning Results Grouped by GAIA Caption Scheme}
\label{tab:captioning}
\centering
\sisetup{table-format=2.2}
\begin{tabular}{@{}lcSSSSS@{}}
\toprule
\multirow{2}{*}{Target} & \centering\multirow{2}{*}{Mode} & 
\multicolumn{5}{c}{Evaluation Metrics (\%)} \\
\cmidrule(lr){3-7}
& & {BLEU} & {ROUGE-1} & {ROUGE-2} & {ROUGE-L} & {METEOR} \\
\midrule

\multirow{2}{*}{Alt-Text} 
& Zero-shot & 0.30 & 14.06 & 2.95 & 12.94 & 10.94 \\
& Fine-tuned & 4.81 & 23.78 & 10.77 & 23.09 & 22.09 \\
\addlinespace[0.15cm]

\multirow{2}{*}{Synthetic}
& Zero-shot & 0.23 & 21.85 & 5.51 & 17.96 & 9.15 \\
& Fine-tuned & 25.62 & 41.48 & 19.30 & 33.35 & 36.08 \\
\bottomrule
\end{tabular}
\vspace{0.2cm}
\end{table*}

\begin{table*}[htbp]
\caption{Remote Sensing Image Captioning (BLIP2) and Cross-Modal Retrieval Results (CLIP-ViT-L-14) on external datasets RSICap and GIT-10M}
\label{tab:captioning_retrieval_combined}
\centering
\scriptsize

% ---- Subtable: Captioning ----
\setlength{\tabcolsep}{4pt} % same spacing for both tables
\sisetup{table-format=2.2}
\begin{tabular}{@{}lcSSSSS@{}}
\toprule
\multirow{2}{*}{Dataset} & \multirow{2}{*}{\centering Mode} & 
\multicolumn{5}{c}{Evaluation Metrics (\%)} \\
\cmidrule(lr){3-7}
& & {BLEU} & {ROUGE-1} & {ROUGE-2} & {ROUGE-L} & {METEOR} \\
\midrule
\multirow{2}{*}{RSICap} 
& Zero-shot & 0.03 & 18.31 & 4.33 & 15.21 & 7.35 \\
& GAIA (Synthetic) & 1.29 & 24.31 & 5.41 & 18.76 & 15.09 \\
\addlinespace[0.15cm]
\multirow{2}{*}{GIT-10M} 
& Zero-shot & 0.05 & 17.41 & 3.40 & 13.95 & 6.83 \\
& GAIA (Synthetic) & 1.16 & 26.75 & 4.37 & 18.15 & 16.62 \\
\bottomrule
\end{tabular}

\vspace{0.4cm}

% ---- Subtable: Retrieval ----
\sisetup{table-format=2.2, table-number-alignment=center}
\begin{tabular}{@{}lc ccccccccc ccccc@{}}
\toprule
\multirow{2}{*}{Dataset} & \multirow{2}{*}{\centering Mode} & 
\multicolumn{5}{c}{Text-to-Image Recall (\%)} & 
\multicolumn{5}{c}{Image-to-Text Recall (\%)} \\
\cmidrule(lr){3-7} \cmidrule(lr){8-12}
& & @5 & @10 & @20 & @50 & @100 & @5 & @10 & @20 & @50 & @100 \\
\midrule
\multirow{2}{*}{RSICap} 
& Zero-shot & 9.79 & 15.67 & 23.91 & 38.99 & 53.93 & 10.44 & 17.02 & 25.65 & 39.65 & 54.16 \\
& GAIA (Synthetic) & 11.26 & 17.06 & 25.80 & 42.40 & 58.18 & 12.83 & 18.29 & 26.72 & 40.68 & 56.00 \\
\midrule
\multirow{2}{*}{GIT-10M} 
& Zero-shot & 1.83 & 2.85 & 4.39 & 7.66 & 11.66 & 2.95 & 4.54 & 6.81 & 11.42 & 16.60 \\
& GAIA (Synthetic) & 4.65 & 6.16 & 8.38 & 12.98 & 17.95 & 5.12 & 6.72 & 9.02 & 13.53 & 18.48 \\
\bottomrule
\end{tabular}

\end{table*}

\begin{table*}[htbp]
\caption{CLIP-ViT-L-14 Classification Performance Comparison between Zero-shot CLIP and RS-domain fine-tuned CLIP models.}
\label{tab:rs_clip_comparison_cls}
\centering
\scriptsize
\sisetup{table-format=2.2, table-number-alignment=center}
\begin{tabular}{@{}l l l SSS@{}}
\toprule
\multicolumn{1}{c}{\multirow{2}{*}{\centering Dataset}} &
\multicolumn{1}{c}{\multirow{2}{*}{\centering Variant}} &
\multicolumn{1}{c}{\multirow{2}{*}{\centering Dataset Size}} &
\multicolumn{3}{c}{Metrics (\%)} \\
\cmidrule(lr){4-6}
& & & {Acc1} & {Acc5} & {Mean Recall} \\
\midrule

\multirow{7}{*}{\centering\rotatebox[origin=c]{90}{eurosat}}
& CLIP                 & -           & 62.57 & 95.93 & 63.95 \\
\cmidrule(lr){2-6}
& GAIA (Alt-text)      & 40K         & 68.52 & 98.87 & 69.75 \\
& GAIA (Synthetic)     & 40K         & 72.46 & 98.91 & 72.68 \\
\cmidrule(lr){2-6}
& SkyCLIP              & 2.6M        & 70.54 & 99.41 & 69.88 \\
& RemoteCLIP           & 165K        & 56.22 & 93.11 & 55.17 \\
& GeoRSCLIP            & 5M          & 74.52 & 99.33 & 75.07 \\
& LAION\_RS            & 726K        & 71.33 & 99.37 & 72.66 \\
\midrule

\multirow{7}{*}{\centering\rotatebox[origin=c]{90}{resisc45}}
& CLIP                 & -           & 63.35 & 92.40 & 63.82 \\
\cmidrule(lr){2-6}
& GAIA (Alt-text)      & 40K         & 65.49 & 91.95 & 66.01 \\
& GAIA (Synthetic)     & 40K         & 66.11 & 92.65 & 66.62 \\
\cmidrule(lr){2-6}
& SkyCLIP              & 2.6M        & 70.59 & 95.19 & 71.23 \\
& RemoteCLIP           & 165K        & 80.57 & 97.87 & 80.82 \\
& GeoRSCLIP            & 5M          & 73.59 & 97.89 & 74.19 \\
& LAION\_RS            & 726K        & 70.11 & 95.18 & 70.63 \\
\midrule

\multirow{7}{*}{\centering\rotatebox[origin=c]{90}{imagenet1k}}
& CLIP                 & -           & 75.54 & 94.58 & 75.54 \\
\cmidrule(lr){2-6}
& GAIA (Alt-text)      & 40K         & 75.32 & 94.66 & 75.33 \\
& GAIA (Synthetic)     & 40K         & 75.49 & 94.57 & 75.48 \\
\cmidrule(lr){2-6}
& SkyCLIP              & 2.6M        & 75.56 & 94.68 & 75.55 \\
& RemoteCLIP           & 165K        & 64.68 & 88.15 & 64.68 \\
& GeoRSCLIP            & 5M          & 66.01 & 90.17 & 66.01 \\
& LAION\_RS            & 726K        & 75.61 & 94.75 & 75.62 \\
\bottomrule
\end{tabular}
\end{table*}

\begin{table*}[htbp]
\caption{CLIP-ViT-L-14 Cross-Modal Retrieval Performance Comparison between Zero-shot CLIP and RS-domain fine-tuned CLIP models using GAIA's thematic tags.}
\label{tab:rs_clip_comparison_retrieval}
\centering
\scriptsize
\setlength{\tabcolsep}{6pt}
\begin{tabular}{@{} l c c c c c c c c c c @{}}
\toprule
\multirow{2}{*}{Model} &
\multicolumn{5}{c}{Text-to-Image Recall (\%)} &
\multicolumn{5}{c}{Image-to-Text Recall (\%)} \\
\cmidrule(lr){2-6} \cmidrule(lr){7-11}
& @5 & @10 & @20 & @50 & @100 & @5 & @10 & @20 & @50 & @100 \\
\midrule
CLIP & 5.51 & 9.17 & 15.09 & 26.60 & 38.07 & 7.41 & 11.52 & 17.05 & 27.55 & 36.97 \\
\midrule
GAIA (Alt-text) & 5.50 & 9.67 & 16.45 & 29.56 & 41.78 & 11.42 & 17.66 & 25.52 & 38.65 & 49.44 \\
GAIA (Synthetic) & 7.76 & 13.04 & 20.72 & 34.28 & 47.32 & 12.30 & 19.07 & 27.30 & 40.41 & 51.79 \\
\midrule
SkyCLIP & 7.67 & 12.35 & 18.85 & 31.26 & 43.18 & 6.74 & 11.06 & 16.20 & 25.92 & 35.80 \\
RemoteCLIP & 1.34 & 2.19 & 3.30 & 6.16 & 9.26 & 3.84 & 6.09 & 9.19 & 16.76 & 24.71 \\
GeoRSCLIP & 4.36 & 7.31 & 10.92 & 19.30 & 28.38 & 5.37 & 8.82 & 13.50 & 22.72 & 32.13 \\
LAION\_RS & 7.74 & 12.99 & 20.46 & 33.08 & 45.03 & 8.90 & 13.77 & 20.07 & 30.76 & 41.19 \\
\bottomrule
\end{tabular}
\end{table*}

\section{Discussion and Conclusion}
In this paper, we introduced GAIA, a novel global, multi-modal, multi-scale vision-language dataset for RS image analysis. GAIA addresses the crucial need for high-quality, domain-specific data to advance the development and application of VLMs in RS. Through a carefully designed data acquisition and annotation pipeline, leveraging web-scraping and the advanced language generation capabilities of GPT-4o, we have created a dataset of remarkable scale, diversity, and semantic richness within the RS domain. Our exploratory data analysis and experimental results robustly demonstrate GAIA's unique characteristics and effectiveness in improving the performance of VLMs on RS tasks. The dataset's global coverage, multi-modal nature, and focus on diverse Earth events, coupled with its scientifically grounded synthetic captions, make GAIA a valuable resource for the RS community.

The release of the GAIA dataset, alongside the automated processing framework and pre-trained model weights, aims to democratize access to high-quality RS vision-language data and accelerate research in this rapidly evolving field. We believe that GAIA will pave the way for the development of more robust, generalizable, and impactful VLMs for RS, enabling new approaches to Earth observation data analysis and unlocking deeper insights into our planet's dynamic systems. Future work will focus on expanding the dataset's scale and thematic scope, further refining annotation quality, exploring its utility across a wider range of VLM architectures and downstream applications, and ultimately, leveraging GAIA to advance the development of next-generation foundation models for RS.

\subsection{Bridging the gap between Remote Sensing and Vision-Language Models}
Our investigation into the current landscape of VLMs and RS datasets revealed a critical gap: existing VLMs, predominantly trained on generic web data, exhibit limited proficiency in the specialized domain of RS, and publicly available RS datasets often lack the scale, diversity, and annotation quality necessary to bridge this gap. GAIA directly addresses these shortcomings through a meticulous two-stage construction process, combining targeted web-scraping from reputable RS sources with advanced caption generation using GPT-4o. This approach has yielded a dataset of 201,005 high-quality image-text pairs, characterized by its global, multi-modal, and multi-scale nature, and importantly, its focus on capturing diverse Earth events and phenomena.

GAIA exhibits a balanced spatial and temporal distribution, spanning 25 years of Earth observations across diverse geographic regions. The inclusion of imagery from over 120 satellite missions and a variety of RS modalities, predominantly optical but also incorporating thermal, radar, and other spectral data, ensures a rich and heterogeneous visual resource. The predominance of moderate-resolution imagery, supplemented by high and very-high resolution data, provides versatility for a wide range of RS analysis tasks, from broad area monitoring to detailed feature extraction. Furthermore, the semantic analysis of GAIA's captions, visualized through word clouds and thematic categorizations, underscores the dataset's focus on dynamic Earth processes and environmental phenomena, encompassing all the five major Earth spheres: Atmosphere, Hydrosphere, Geosphere, Biosphere, and Cryosphere. The quantitative comparison of synthetic captions with original alt-text highlights the enhancement in textual richness and detail achieved through our annotation pipeline. Synthetic captions exhibit a substantially larger vocabulary, greater length, and more complex sentence structures, offering a far more comprehensive and nuanced textual grounding for the visual content.

Our experiments demonstrate GAIA’s value across vision-language tasks. Fine-tuning CLIP on GAIA improved RS image classification and cross-modal retrieval performance across all model scales (ViT-B-32, ViT-L-14, ViT-L-14-336), with synthetic captions outperforming alt-text due to their enhanced semantic richness. Gains were most pronounced on EuroSAT (addressing zero-shot CLIP’s RS limitations), while RESISC45 saw more modest improvements, a disparity potentially tied to GAIA’s thematic focus and limited high-resolution imagery. Crucially, GAIA fine-tuning preserved generalizability, maintaining or slightly improving ImageNet1K performance, which underscores positive transfer learning. Similarly, BLIP2 fine-tuned on GAIA achieved substantial gains in captioning metrics and generated more contextually accurate descriptions of Earth events, though limitations like geographic inaccuracies and verbosity persisted. Together, these results highlight GAIA’s dual role: enhancing RS-specific model capabilities through domain-adapted training while retaining foundational visual understanding, with captioning challenges pointing to opportunities for refining annotation pipelines and model architectures.

\subsection{Future Directions for GAIA and Remote Sensing Vision-Language Models}
While the initial experimental results are promising, it is crucial to acknowledge inherent limitations within the scope of this study. Although GAIA represents a substantial advancement in RS dataset scale and annotation quality, its construction is inherently bounded by the availability of publicly accessible RS imagery and associated textual metadata online. Furthermore, the dataset's thematic emphasis, while advantageous for numerous Earth science applications, may also introduce inherent biases, potentially limiting its generalizability to a broader spectrum of RS tasks beyond event-centric scenarios. It is also crucial to note that the synthetic captions are designed to augment the semantic context provided by the original alt-text, rather than to function as exhaustive, dense annotations capturing every discrete visual element. 

To systematically address these identified limitations, our future research trajectory encompasses a series of follow-up investigations. Given the rapid advancements in multi-modal large language models (MLLMs) and the increasing convergence in performance between open-source, open-weight and proprietary models, we strongly believe that it will be feasible to scale our annotation pipeline in a cost-effective manner, enabling us to integrate more data sources. This expansion will involve both direct web-scraping from the internet and the incorporation of existing web-crawled datasets, such as the filtered RS image-text pairs referenced in Section~\ref{sec:seen_rs}. Such an approach will enrich GAIA with more generic RS information, moving beyond a predominantly event-oriented sample distribution, and concurrently address the under-representation of complementary data often found in existing datasets such as aerial imagery and very high spatial resolution imagery in general. Moreover, the inclusion of additional dense captions, alongside a dedicated instruction-tuning dataset, is considered essential for the effective training and fine-tuning of state-of-the-art models for the specific challenges of RS vision-language tasks. This is particularly relevant as the preliminary experiments presented herein are limited in scope, primarily evaluating CLIP and BLIP2 models across a subset of common RS applications. Further research is therefore needed to explore the full potential of GAIA across a wider range of VLM architectures and diverse downstream applications within the RS domain. Last but not least, beyond visual and textual data, a critical enhancement for GAIA will be the integration of corresponding RS modalities, such as multi-spectral and Synthetic Aperture Radar (SAR) data. This will offer crucial, complementary information, leading to more comprehensive scene understanding and improved robustness in adverse conditions. Multi-modal VLM approaches have recently seen increased interest \cite{zavras2024mind,khanal2023soundscape,klemmer2023satclip,vivanco2024geoclip} in RS. This dataset expansion is expected to significantly enhance the capabilities of models trained on GAIA and enable the RS community to explore the full potential of our dataset across a wider range of VLM architectures and diverse downstream applications within the RS domain.

\section*{Acknowledgments}
This work has received funding from the European Union's Horizon Europe research and innovation project ThinkingEarth under grant agreement number 101130544. This work was developed during the research stay of Angelos Zavras at the Remote Sensing Image Analysis (RSiM) Group of the Faculty of Electrical Engineering and Computer Science, Technische Universität Berlin. The research stay grant was awarded by the Short-Term Research Grants program (57693450) of the German Academic Exchange Service (DAAD).

\bibliographystyle{IEEEtran}
\bibliography{main.bib}

\end{document}